\documentclass{article}
\usepackage{iclr2026_conference,times}
\iclrfinalcopy

\usepackage[dvipsnames]{xcolor}
\definecolor{citecolor}{HTML}{1520a6}
\definecolor{linkcolor}{HTML}{900603}
\definecolor{bordo}{HTML}{ff0040}

\usepackage{hyperref}
\hypersetup{
  colorlinks=true,
  citecolor=citecolor,
  linkcolor=linkcolor,
  urlcolor=black,
}

\usepackage{graphicx}
\usepackage{subcaption}
\usepackage{mathtools}
\usepackage{microtype}
\usepackage{float}
\usepackage{url}
\usepackage{booktabs}
\usepackage{enumitem}
\usepackage{amsthm}
\usepackage{amsmath}
\usepackage{amssymb}
\usepackage{tikz}
\usepackage{newtxmath}
\usepackage{wrapfig}
\usepackage{booktabs}
\usepackage{xcolor}
\usepackage{fontawesome5}
\usepackage[linesnumbered,ruled,vlined]{algorithm2e}

\usepackage{cleveref}

\SetCommentSty{mycommfont}

\newcommand{\std}[1]{\scriptsize $\pm$#1}
\newcommand{\ie}{\textit{i.e.}}
\newcommand{\eg}{\textit{e.g.}}
\DeclareMathOperator*{\argmax}{argmax}
\DeclareMathOperator*{\argmin}{argmin}

\title{
  Data-to-Energy Stochastic Dynamics
}

\author{Kirill Tamogashev \\
  University of Edinburgh \\
  \texttt{k.tamogashev@sms.ed.ac.uk}
  \And
  Esmeralda S. Whitammer \\
  University of Edinburgh, CIFAR Fellow \\
  \texttt{esmeralda.whitammer@ed.ac.uk} 
  \AND
  \hspace{32mm} \href{https://github.com/mmacosha/d2e-stochastic-dynamics}{\textcolor{citecolor}{
  \faGithub \; \texttt{mmacosha/d2e-stochastic-dynamics}}} 
}

\makeatletter
\crefname{section}{\S\@gobble}{\S\@gobble}
\crefname{subsection}{\S\@gobble}{\S\@gobble}
\crefname{proposition}{Prop.}{Props.}
\crefname{figure}{Fig.}{Figs.}
\crefname{table}{Table}{Tables}
\renewcommand{\eqref}[1]{(\ref{#1})}

\def\paragraph{\@startsection{paragraph}{4}{\z@}{0ex}{-1em}{\normalsize\bf}}

\makeatother

\begin{document}

\maketitle

\begin{abstract}
  \looseness=-1
  The Schrödinger bridge problem is concerned with finding a
  stochastic dynamical system bridging two marginal distributions
  that minimises a certain transportation cost.
  This problem, which represents a generalisation of optimal
  transport to the stochastic case, has received attention due to its
  connections to diffusion models and flow matching, as well as its
  applications in the natural sciences.
  However, all existing algorithms enable the inference of such
  dynamics only for cases where samples from both distributions are available.
  In this paper, we propose the first general method for modelling
  Schrödinger bridges when one (or both) distributions are given by
  their unnormalised densities, with no access to data samples.
  Our algorithm relies on a generalisation of the iterative
  proportional fitting (IPF) procedure to the data-free case,
  inspired by recent developments in off-policy reinforcement
  learning for training of diffusion samplers.
  We demonstrate the efficacy of the proposed \emph{data-to-energy
  IPF} on synthetic problems, finding that it can successfully learn
  transports between multimodal distributions.
  As a secondary consequence of our reinforcement learning
  formulation, which assumes a fixed time discretisation scheme for
  the dynamics, we find that existing data-to-data Schrödinger bridge
  algorithms can be substantially improved by learning the diffusion
  coefficient of the dynamics.
  Finally, we apply the newly developed algorithm to the problem of
  sampling posterior distributions in latent spaces of generative
  models, thus creating a data-free image-to-image translation method.
\end{abstract}

\section{Introduction}

Two modern approaches to generative modelling that have paved the way
for scalable and efficient generation of high-fidelity images
\citep{dhariwal2021diffusion, rombach2021stablediffusion}, videos
\citep{polyak2024moviegen}, audio \citep{chen2021wavegrad,
kong2021diffwave} and text \citep{nie2025lldm,
sahoo2025diffusionduality} are diffusion models and flow matching.
Diffusion models \citep{sohl2015diffusion, ho2020denoising,
song2021scorebased} assume a noising stochastic process that
transforms data into a tractable noise distribution and use
score-based techniques to learn its reverse process, which transforms
noise into data.
Flow matching \citep{liu2022flow, albergo2023stochastic,
lipman2023flow, tong2024improving} learns time-dependent
deterministic dynamics that give a transportation map between two
arbitrary distributions.
Both approaches can be seen as special cases of the more general
problem of learning stochastic dynamics between two arbitrary distributions.

The problem of inferring an optimal stochastic bridge between two
distributions is called the Schrödinger bridge (SB) problem. Initially proposed in \citet{schrodinger1931umkehrung,
schrodinger1932theorie}, it has recently been studied using various
machine learning techniques \citep{huang2021schrodinger,
  vargas2021solving, chen2021likelihood, stromme2023sampling,
shi2023dsbm,tong2024simulation}.
One computational approach to the Schrödinger bridge problem is the
iterative proportional fitting (IPF) algorithm
\citep{fortet1940resolution,vargas2021solving,debortoli2021diffusion},
which maintains a pair of processes in forward and reverse time and
iteratively updates them by solving half-bridge problems (see
\Cref{sec:d2d_ipf}).
Upon convergence, the two processes become time reversals of each
other and solve the SB problem.
Notably, the typical training of diffusion models -- with a fixed
noising process that transforms a data distribution into a Gaussian
by construction -- is a degenerate case of IPF that converges in a single step.
However, existing variants of IPF work only in a setting where
samples from both marginal distributions are available and thus
cannot be used to model bridges when one (or both) of the marginal
distributions is given as an unnormalised density, without access to
data samples.

We propose to extend the IPF algorithm to the case where one or both
marginal distributions are given by unnormalised densities or energy
functions: $p(x) = e^{-\mathcal{E}(x)} / Z,\;Z = \int
e^{-\mathcal{E}(x)}\, {\rm d}x$, where $\mathcal{E}$ can be queried,
but $Z$ is unknown.
Our proposed \emph{data-to-energy} (or \emph{energy-to-energy}) IPF
generalises recently developed techniques for training diffusion
models to sample from a distribution given by an unnormalised density
\citep[][\textit{inter
alia}]{zhang2022path,vargas2023denoising,vargas2024transport,berner2024optimal,albergo2025nets,blessing2025underdamped}.
In particular, we build upon off-policy reinforcement learning losses
and stabilisation techniques for diffusion samplers
\citep{richter2020vargrad,
richter2024improved,lahlou2023theory,sendera2024improved,gritsaev2025adaptive}
to propose an efficient training for the IPF steps in the
data-to-energy setting.
Our algorithm is the first general method for inferring
data-to-energy and energy-to-energy Schrödinger bridges.

Wielding the newly proposed algorithm, we make three contributions:
\begin{enumerate}[left=0pt,nosep,label=(\arabic*)]
  \item We show that the proposed data-to-energy and energy-to-energy
    IPF algorithms successfully learn stochastic bridges with low
    transport cost between synthetic datasets and densities,
    performing on par with the transports learnt by data-to-data IPF
    using samples from a ground-truth oracle.
  \item As a secondary contribution, we show that -- as a consequence
    of the time discretisation used in our reinforcement learning
    formulation -- existing data-to-data IPF algorithms can be
    improved by learning the diffusion coefficient of the dynamics,
    in addition to the drift, generalising the results of
    \citet{gritsaev2025adaptive} for diffusion samplers to the more
    general SB setting.
  \item We apply the data-to-energy IPF algorithm to the problem of
    translating prior distributions to posteriors in latent spaces of
    generative models, generalising the outsourced diffusion sampling
    of \citet{venkatraman2025outsourced} to yield a scalable
    data-free image-to-image translation method.
\end{enumerate}

\section{Data-to-data Schrödinger bridges}
\label{sec:d2d_ipf}

\subsection{Iterative proportional fitting for data-to-data SB}

\paragraph{Setting: The SB problem and its connection to optimal
transport.} We present some background on the SB problem; see
\citet{leonard2013survey,debortoli2021diffusion} for relevant and
more detailed overviews.
Let $p_0$ and $p_1$ be two given distributions over the space
$\mathbb{R}^d$, assumed to be absolutely continuous (thus used
interchangeably with their density functions) and of finite variance.
Let $\mathbb{Q}_t$ be a reference process on the time interval
$[0,1]$ taking values in $\mathbb{R}^d$ (usually an
Ornstein-Uhlenbeck process, such as the Wiener process). The
Schrödinger bridge problem can be formalised as:
\begin{equation}
  \mathbb{P}^{*}_t = \argmin\limits_{\mathbb{P}_t} \left\{
    {\rm KL}\left(\mathbb{P}_t\,\|\, \mathbb{Q}_t\right) \;
    {\rm s.t.\ } (\pi_0)_{\#} \mathbb{P}_t = p_0, (\pi_1)_{\#}
    \mathbb{P}_t = p_1
  \right\},
  \label{eq:sb-problem}
\end{equation}
where the minimisation is taken over all processes $\mathbb{P}_t$
whose marginals at times $t=0$ and $t=1$, written $(\pi_0)_{\#}
\mathbb{P}_t$ and $(\pi_1)_{\#} \mathbb{P}_t$, equal $p_0$ and $p_1$,
respectively. The solution $\mathbb{P}^{*}_t$ is a stochastic
dynamical system that transports $p_0$ to $p_1$ -- that is, a
\emph{bridge} -- that is the closest to $\mathbb{Q}_t$ in KL divergence.
If the reference process $\mathbb{Q}_t$ is given by an It\^o
stochastic differential equation (SDE)
\begin{equation*}
  \mathbb{Q}_t:\quad\mathrm{d}X_t=F_{\rm
  ref}(X_t,t)\,\mathrm{d}t+\sigma_t\,\mathrm{d}W_t,\quad X_0\sim q_0,
\end{equation*}
then, under mild conditions (see \citet{leonard2013survey}) the
solution to \eqref{eq:sb-problem} exists, is unique, and also takes
the form of an SDE:
\begin{equation*}
  \mathbb{P}_t:\quad\mathrm{d}X_t=F(X_t,t)\,\mathrm{d}t+\sigma_t\,\mathrm{d}W_t,\quad
  X_0\sim p_0.
\end{equation*}
with the same diffusion coefficient. The KL then takes the form of a
dynamic transport cost \citep{berner2025discrete}
\begin{equation}
  {\rm KL}(\mathbb{P}_t\,\|\,\mathbb{Q}_t)
  =
  {\rm KL}(p_0\,\|\,q_0) +
  \mathbb{E}_{X_t\sim\mathbb{P}_t}\int_0^1\frac{\|F_{\rm
  ref}(X_t,t)-F(X_t,t)\|^2}{2\sigma_t^2}\,\mathrm{d}t,
  \label{eq:cost}
\end{equation}
reducing the problem \eqref{eq:sb-problem} to one of inferring the
drift function $F$ minimising the cost \eqref{eq:cost}. This
representation makes clear that as $\sigma_t\to 0$, the SB problem
approaches the dynamic optimal transport problem between $p_0$ and
$p_1$ with squared-euclidean cost (see \citet{tong2024improving}).
For $\sigma_t>0$, the joint marginal distribution of $\mathbb{P}_t^*$
over $X_0,X_1$ is an \emph{entropy-regularised} optimal transport, a
key observation in the derivation of the SB algorithm in
\citet{tong2024simulation}.

\paragraph{Iterative proportional fitting.}
Computationally, the SB problem can be solved using the iterative
proportional fitting (IPF) algorithm \citep{fortet1940resolution,
vargas2021solving,debortoli2021diffusion}. IPF defines a recursion
initialised at $\overrightarrow{\mathbb{P}}^0_t = \mathbb{Q}_t$:
\begin{subequations}\label{eq:ipf-iterations}
  \begin{align}
    \overleftarrow{\mathbb{P}}^{n+1}_t
    &=\argmin\limits_{\mathbb{P}_t} \Big\{
      {\rm KL}\big(\mathbb{P}_t\,\|\,\overrightarrow{\mathbb{P}}^{n}_t\big) \,
      \text{s.t.}\, (\pi_0)_{\#}\mathbb{P}_t = p_0
    \Big\}
    &&= p_0\otimes\overrightarrow{\mathbb{P}}^{n}_{t|0}, \label{eq:ipf-bwd}
    \\
    \overrightarrow{\mathbb{P}}^{n+1}_t
    &= \argmin\limits_{\mathbb{P}_t} \Big\{
      {\rm KL}\big(\mathbb{P}_t\,\|\,\overleftarrow{\mathbb{P}}^{n+1}_t\big) \,
      \text{s.t.}\, (\pi_1)_{\#}\mathbb{P}_t = p_1
    \Big\}
    &&= p_1\otimes\overleftarrow{\mathbb{P}}^{n+1}_{t|1}, \label{eq:ipf-fwd}
  \end{align}
\end{subequations}
where each step is the solution to a \emph{half-bridge} problem,
which pins the previous iterate at one of the marginals $p_0$ or
$p_1$ (here $\overrightarrow{\mathbb{P}}^{n}_{t|0}$ denotes the
  conditional process given $X_0$ and
  $\overleftarrow{\mathbb{P}}^{n+1}_{t|1}$ the conditional process
given $X_1$). IPF for Schrödinger bridges is thus a dynamic
generalisation of the Sinkhorn algorithm
\citep{sinkhorn1964relationship}, which computes entropic optimal
transport by iteratively renormalising a cost matrix over rows and columns.
It can be shown \citep{debortoli2021diffusion} that the iterates
$\overrightarrow{\mathbb{P}}_t^n$ and
$\overleftarrow{\mathbb{P}}_t^n$ converge to the same stochastic
process and that this process solves the SB problem \eqref{eq:sb-problem}.

If $\overrightarrow{\mathbb{P}}^0_t=\mathbb{Q}_t$ is represented as
an SDE, then the iterates defined in \eqref{eq:ipf-fwd} and
\eqref{eq:ipf-bwd} can be represented as forward-time and
reverse-time SDEs initialised at $p_0$ and $p_1$, respectively. We
can thus introduce SDEs with neurally parametrised drifts
\citep{tzen2019neural}:
\begin{subequations}
  \label{eq:bwd-fwd-sdes}
  \begin{align}
    \overrightarrow{\mathbb{P}}_t^n:\quad&{\rm d}\overrightarrow{X}_t
    = \overrightarrow{F}_{\theta_n}(X_t, t) \,{\rm d}t
    + \sigma_t\,{\rm d}\overrightarrow{W}_t, \label{eq:fwd-sde}
    \\
    \overleftarrow{\mathbb{P}}_t^n:\quad&{\rm d}\overleftarrow{X}_t
    = \overleftarrow{F}_{\varphi_n}(X_t, t)\,{\rm d}t
    + \sigma_t\,{\rm d}\overleftarrow{W}_t, \label{eq:bwd-sde}
  \end{align}
\end{subequations}
where $\sigma_t$ coincides with the diffusion coefficient of the
reference process, and perform the iterations \eqref{eq:ipf-iterations}
as optimisation problems over $\theta_n$ and $\varphi_n$.
(We henceforth occasionally drop the subscript $n$ from the parameters,
  with the understanding that \eqref{eq:ipf-bwd} optimises
  $\varphi=\varphi_{n+1}$ under a fixed $\theta_n$ and \eqref{eq:ipf-fwd}
optimises $\theta=\theta_{n+1}$ under a fixed $\varphi_{n+1}$.)

\paragraph{Data-to-data IPF in a time discretisation.}
To approximately perform the optimisations involved in IPF with
respect to the parameters $\theta$ and $\varphi$, we discretise the
SDEs \eqref{eq:bwd-fwd-sdes} representing
$\overrightarrow{\mathbb{P}}_t$ and  $\overleftarrow{\mathbb{P}}_t$
over $K$ steps using the Euler-Maruyama scheme with $\Delta
t=\frac1K$. The continuous-time processes are thus approximated by
discrete-time Markov chains, \ie, joint distributions over
discrete-time trajectories $\tau=(x_0,x_{\Delta t}\dots,x_1)$, having
the factorisation:
\begin{subequations}
  \small
  \label{eq:discretisation}
  \begin{align}
    \!\!\!\!\!\!\overrightarrow{p}_{\theta}(\tau)
    &\!=\! p_0(x_0) \overbracket{\prod^{K - 1}_{k=0}
      \overrightarrow{p}_{\theta}(x_{(k + 1)\Delta t} \!\mid\!
    x_{k\Delta t})}^{\overrightarrow{p}_\theta(\tau\mid x_0)}, \;
    \overbracket{\overrightarrow{p}_{\theta}(x_{(k + 1)\Delta t}
      \!\mid\! x_{k\Delta t} )
      \!=\! \mathcal{N}\left(
        x_{k\Delta t} + \overrightarrow{F}_{\theta}( x_{k\Delta t},
        k\Delta t) \Delta t,
        \sigma_{k\Delta t}^2 \Delta t
    \right)}^{\text{Euler-Maruyama step with step size $\Delta t$}},
    \label{eq:fwd-param}
    \\
    \!\!\!\!\!\!\overleftarrow{p}_{\varphi}(\tau)
    &\!=\! p_1(x_1) \underbracket{\prod^K_{k=1}
      \overleftarrow{p}_{\varphi}(x_{(k - 1)\Delta t} \!\mid\!
    x_{k\Delta t} )}^{\overleftarrow{p}_\varphi(\tau\mid x_1)}, \;
    \underbracket{\overleftarrow{p}_{\varphi}(x_{(k - 1)\Delta t}
      \!\mid\! x_{k\Delta t} )
      \!=\! \mathcal{N}\left(
        x_{k\Delta t} + \overleftarrow{F}_{\varphi}( x_{k\Delta t},
        k\Delta t) \Delta t,
        \sigma_{k\Delta t}^2 \Delta t
    \right)}_{\text{reverse-time Euler-Maruyama step with step size
    $\Delta t$}}. \label{eq:bwd-param}
  \end{align}
\end{subequations}
As proposed in \citet{vargas2021solving}, the two IPF optimisation
problems in \eqref{eq:ipf-iterations} can be approximately solved by
maximum likelihood. On the level of the discretised processes, this
amounts to the following recurrence:
\begin{subequations}\label{eq:tlm-iterations}
  \begin{align}
    \varphi_{n+1} &= \argmax\limits_{\varphi}
    \mathbb{E}_{x_0\sim p_0,\tau \sim
    \overrightarrow{p}_{\theta_{n}}(\tau\mid x_0)}
    \sum_{k=1}^{K} \log \overleftarrow{p}_{\varphi}(x_{(k - 1)\Delta
    t} \mid x_{k\Delta t} )%
    \label{eq:tlm-bwd}
    \\
    \theta_{n+1} &= \argmax\limits_{\theta}
    \mathbb{E}_{x_1\sim p_1,\tau \sim
    \overleftarrow{p}_{\varphi_{n+1}}(\tau\mid x_1) }
    \sum_{k=0}^{K-1} \log \overrightarrow{p}_{\theta}(x_{(k+1)\Delta
    t} \mid x_{k\Delta t})
    \label{eq:tlm-fwd}
    .
  \end{align}
\end{subequations}
The optimisation problem in \eqref{eq:tlm-bwd}
(resp.\ \eqref{eq:tlm-fwd}) requires taking a sample from the
marginal distribution $x_0\sim p_0$ (resp.\ $x_1\sim p_1$), rolling
out a trajectory in forward time from $\overrightarrow{p}_\theta$
(resp.\ in reverse time from $\overleftarrow{p}_\varphi$) initialised
at the sample, and maximising the log-likelihood of this trajectory
in the opposite direction, \ie, under $\overleftarrow{p}_\varphi$
(resp.\ under $\overrightarrow{p}_\theta$). This procedure
essentially requires samples from $p_0$ (resp.\ from $p_1$) to be available.

We call this iterative algorithm the \emph{log-likelihood method} and
give an algorithmic presentation in \Cref{alg:d2d-ipf}. It contrasts
with the method proposed in \citet{debortoli2021diffusion}, which
uses a slightly different discretisation scheme, although the two
coincide in the continuous-time limit ($K\to\infty$).

\paragraph{Diffusion models as a special case.}

Diffusion models \citep{sohl2015diffusion, ho2020denoising,
song2021scorebased} can be seen as a special case of the IPF
algorithm that converges in a single step. If $p_0$ is a data
distribution and $p_1$ is Gaussian, and $\mathbb{Q}_t$ is a noising
process that transports any source distribution to $p_1$ by
construction, then $\mathbb{Q}_t$ initialised at $p_0$ is already a
bridge between $p_0$ and $p_1$. Thus the first iteration of IPF --
learning $\overleftarrow{\mathbb{P}}_t^1$ by maximum-likelihood
training \eqref{eq:tlm-bwd} on trajectories sampled from
$p_0\otimes\mathbb{Q}_t|_0$ -- already yields a bridge between $p_0$
and $p_1$, so all subsequent iterations are redundant and
$\overleftarrow{\mathbb{P}}_t^1$ solves the SB problem. In practice,
training $\overleftarrow{\mathbb{P}}_t^1$ as a neural SDE proceeds by
score matching, which is simply a Rao-Blackwellised estimate of the
IPF maximum-likelihood objective \citep{song2021maximum}.

\subsection{Discretisation allows flexible kernels}

The majority of existing work
\citep{chen2021likelihood, vargas2021solving, debortoli2021diffusion,
shi2023dsbm}
trains only the drift functions
$\overrightarrow{F}_\theta,\overleftarrow{F}_\varphi$
or related objects using objectives similar to \eqref{eq:tlm-iterations}.
In contrast to that, \textbf{we propose to train not only the drift,
but also diffusion coefficients of both processes}, by replacing the
variances $\sigma^2_{k\Delta t}$ in \eqref{eq:discretisation} by
learnt functions
$\overrightarrow{\sigma}^2_\theta(x_k,k\Delta t)$ and
$\overleftarrow{\sigma}^2_\varphi(x_k,k\Delta t)$.
We expect this to correct for the effect of time discretisation
error, inspired by the results
for diffusion samplers in \citet{gritsaev2025adaptive}.

\looseness=-1
The optimisation problems in \eqref{eq:tlm-iterations} can then be
solved with respect to the parameters of both the drift and diffusion
coefficients.
We compare this approach to those that do not learn the variance in
\Cref{tab:comparison_d2d}.

\section{Data-free Schrödinger bridges}

\begin{figure}[t]
  \begin{minipage}[t]{0.48\textwidth}
    \begin{algorithm}[H]
      \caption{Data-to-Data IPF}
      \label{alg:d2d-ipf}
      \DontPrintSemicolon

      \For{$n = 0, \ldots, n_{\rm max}$}{
        \tcc{Backward IPF step \eqref{eq:tlm-bwd}}
        \While{not converged}{
          $x_0\sim p_0$\;
          $\tau=(x_0,x_{\Delta t},\dots,x_1)\sim
          \overrightarrow{p}_\theta(\tau\mid x_0)$ using \eqref{eq:fwd-param}\;
          Gradient step on $\varphi$ with
          $\nabla\log\overleftarrow{p}_\varphi(\tau\mid x_1)$
        }
        \tcc{Forward IPF step \eqref{eq:tlm-fwd}}
        \While{not converged}{
          $x_1 \sim p_1$\;
          $\tau=(x_0,\dots,x_{1-\Delta t},x_1)\sim
          \overleftarrow{p}_\varphi(\tau\mid x_1)$ using \eqref{eq:bwd-param}\;
          Gradient step on $\theta$ with
          $\nabla\log\overrightarrow{p}_\theta(\tau\mid x_0)$
        }
      }
      \KwRet $\theta$, $\varphi$\;
    \end{algorithm}
  \end{minipage}
  \hfill
  \begin{minipage}[t]{0.48\textwidth}
    \begin{algorithm}[H]
      \caption{Data-to-Energy IPF}
      \label{alg:d2e-ipf}
      \DontPrintSemicolon
      \textcolor{bordo}{
        Initialise buffer $\mathcal{B}=\emptyset$\;
      }
      \For{$n = 0, \ldots, n_{\rm max}$}{
        \tcc{Backward IPF step \eqref{eq:tlm-bwd}}
        \While{not converged}{
          $x_0\sim p_0,\tau\sim \overrightarrow{p}_\theta(\tau\mid x_0)$\;
          Gradient step on $\varphi$ with
          $\nabla\log\overleftarrow{p}_\varphi(\tau\mid x_1)$\;
          \textcolor{bordo}{
            Update $\cal B$ with samples $x_1$
          }
        }

        \tcc{Forward IPF step \eqref{eq:d2e-loss}}
        \While{not converged}{
          \textcolor{bordo}{
            \If{on-policy}{
              $x_0\sim p_0,\tau\sim\overrightarrow{p}_\theta(\tau\mid x_0)$
            }
            \Else{
              $x_1\sim\mathcal{B},\tau^{(1)}\sim
              \overleftarrow{p}_\varphi(\tau\mid x_1)$\;
              $\tau^{(2)},\dots,\tau^{(N)}\sim
              \overrightarrow{p}_\theta(\tau\mid x_0^{(1)})$
            }
            Gradient step on $\theta$ with $\nabla{\rm Var}_i \left(
              \log \frac{\overrightarrow{p}_{\theta}(\tau^{(i)}\mid
              x_0^{(1)})}{\overleftarrow{p}_{\varphi}(\tau^{(i)}\mid
              x_1^{(i)})}+\mathcal{E}_1(x_1^{(i)})
            \right)$
        }}
      }
      \KwRet $\theta$, $\varphi$\;
    \end{algorithm}
  \end{minipage}
  \caption{\textbf{Left:} Algorithm for data-to-data IPF.
    \textbf{Right:} Algorithm for data-to-energy IPF, showing the
    replay buffer with backward trajectory reuse
    (\Cref{sec:d2e_theory}), with differences highlighted in
  \textcolor{bordo}{red}.}
  \label{fig:algos}
\end{figure}

\subsection{IPF for data-to-energy SB}\label{sec:d2e_theory}

We now consider the setting where samples from $p_0$ are available,
but $p_1$ is given by an unnormalised density $p_1(x) =
e^{-\mathcal{E}_1(x)} / Z$ and $Z = \int e^{-\mathcal{E}_1(x)} {\rm d}x$
is unknown. In this case, the odd-numbered IPF steps
\eqref{eq:ipf-bwd} can be performed (via \eqref{eq:tlm-bwd}), but the
even-numbered steps \eqref{eq:ipf-fwd} cannot be done using
\eqref{eq:tlm-fwd}, as they require samples from $p_1$. Instead, we
need an objective that would fit $\overrightarrow{\mathbb{P}}^{n+1}_t$ as a
forward-time SDE matching
$p_1\otimes\overleftarrow{\mathbb{P}}^{n+1}_{t|1}$ without samples from $p_1$.

In the time discretisation \eqref{eq:discretisation}, the IPF step
\eqref{eq:ipf-fwd} for $\overrightarrow{\mathbb{P}}^{n+1}_t$ requires
enforcing that for every $x_0$, $\overrightarrow{p}_{\theta}(\tau\mid
x_0) \propto \overleftarrow{p}_{\varphi}(\tau \mid x_1) p_1(x_1)$
over all trajectories $\tau=(x_0,x_{\Delta t},\dots,x_1)$ starting at $x_0$.

To approximately enforce this proportionality, we introduce a
source-conditional variant of the second-moment, or log-variance,
loss used for training diffusion samplers
(\citet{richter2020vargrad}; see \citet{berner2025discrete} for an
  overview of related losses and their consistency properties in the
continuous-time limit). The idea is to minimise the variance of the
log-ratio of the two sides of the proportionality over trajectories
$\tau$ sharing the same $x_0$:
\begin{equation}
  \mathcal{L}_{\rm LV}(x_0,\theta)
  = {\rm Var} \left(
    \log \frac{\overrightarrow{p}_{\theta}(\tau\mid
    x_0)}{p_1(x_1)\overleftarrow{p}_{\varphi}(\tau\mid x_1)}
  \right)
  ={\rm Var} \left(
    \sum_{k=1}^{K} \log \frac{\overrightarrow{p}_{\theta}(x_{k\Delta
    t} \mid x_{(k - 1)\Delta t} ) }{\overleftarrow{p}_{\varphi}(x_{(k
    - 1)\Delta t} \mid x_{k\Delta t} )} + \mathcal{E}_1(x_1)
  \right),
  \label{eq:vargrad}
\end{equation}
where the variance is taken over some full-support distribution
$p^{\rm train}(\cdot\mid x_0)$ over trajectories. (The normalising
  constant $Z$ does not affect the variance, nor do we need to know the
marginal of the process being learnt at time 0.) The empirical
variance over a batch of $N$ trajectories sampled from $p^{\rm
train}(\tau\mid x_0)$ is an unbiased estimate of \eqref{eq:vargrad}
and can be used in the loss. (The training policy we use below takes
  $N$ \emph{non-i.i.d.}\ trajectories, thus $p^{\rm train}(\cdot\mid
x_0)$ can be a distribution over \emph{batches} of $\tau$.)

Because this proportionality must hold for every $x_0$, we must also
average \eqref{eq:vargrad} over $x_0\sim p_0^{\rm train}(x_0)$, where
$p_0^{\rm train}$ is some training distribution over $x_0$. Thus the
full objective for the forward IPF step is
\begin{equation}
  \mathcal{L}(\theta) = \mathbb{E}_{x_0\sim p_0^{\rm train}(x_0)}
  {\mathbb E}_{\tau^{(1)},\dots,\tau^{(N)}\sim p^{\rm
  train}(\cdot\mid x_0)}\left[{\rm Var}_i \left(
      \log \frac{\overrightarrow{p}_{\theta}(\tau^{(i)}\mid
      x_0)}{\overleftarrow{p}_{\varphi}(\tau^{(i)}\mid
      x_1^{(i)})}+\mathcal{E}_1(x_1^{(i)})
  \right) \right].
  \label{eq:d2e-loss}
\end{equation}
The choice of $p^{\rm train}_0(x_0)$ and $p^{\rm train}(\cdot\mid
x_0)$ is very important and we discuss it in the next subsection.

\paragraph{Comparison with diffusion samplers.}
The objective \eqref{eq:vargrad} is a conditional variant of the
log-variance, or \emph{VarGrad}, loss \citep{richter2020vargrad}
previously used for diffusion-based samplers of unnormalised target
densities. In that setting -- \emph{but not in ours}, since the
marginal of $p_1\otimes\overleftarrow{\mathbb{P}}^{n+1}_{t|1}$ at
time 0 is not $p_0$ unless IPF has converged -- the density of the
process being learnt at time 0 is known to be $p_0(x_0)$. Thus
$p_0(x_0)$ can be placed in the numerator of the loss in
\eqref{eq:vargrad} and variance can be taken over both $x_0$ and
$\tau$. An alternative objective would fit the density at $x_0$,
yielding a variant of the related \emph{trajectory balance} (TB) loss
for diffusion samplers \citep{malkin2022trajectory,lahlou2023theory}.

\looseness=-1
Yet another alternative would avoid IPF altogether and simply perform
joint optimisation of both $\theta$ and $\varphi$ using a VarGrad- or
TB-like objective to enforce that
$p_0(x_0)\overrightarrow{p}_\theta(\tau\mid x_0) =
p_1(x_1)\overleftarrow{p}_\varphi(\tau\mid x_1)$ over all
trajectories $\tau$. Such a \emph{bridge sampling} approach is taken
in \citet{blessing2025underdamped, blessing2025endtoend,
gritsaev2025adaptive}. However, this approach would yield a bridge
that is not necessarily the solution to the SB problem
\eqref{eq:sb-problem}, since the KL to the reference process is not
minimised. In a future work, it would be interesting to compare our
IPF approach with those that regularise bridge sampling losses by the
cost \eqref{eq:cost}.

\subsection{Off-policy training methods} \label{sec:off-policy}
The objective \eqref{eq:d2e-loss} leaves room for the choice of
training distributions
$p_0^{\rm train}(x_0)$ and $p^{\rm train}(\tau\mid x_0)$, which can
vary over the course of training. In this way, training with this
loss is a form of off-policy reinforcement learning, a connection
that has been elucidated and exploited for improved training of
diffusion samplers in \citet{sendera2024improved}.

A na\"ive choice would take
$p_0^{\rm train}=p_0$ and
$p^{\rm train}(\tau\mid x_0)=\overrightarrow{p}_\theta(\tau\mid x_0)$
(\emph{on-policy} training).
However, for complex, high-dimensional distributions on-policy
training is insufficient, as modes that are not discovered by the
sampler are very unlikely to be explored. To facilitate
training we adapt practices from diffusion sampling literature to
guide the sampling
process towards the areas of high density of the target distribution
$p_1$. In the following paragraphs, we discuss such techniques.

\paragraph{Replay buffer.} We keep a replay buffer of final samples
$x_1$ from the process $\overrightarrow{p}_{\theta}(\tau \mid x_0)$.
During training, to obtain $x_0$, we sample $x_1$ from the buffer,
then sample a reverse trajectory to obtain $x_0$ for training:
$\tilde{x}_0 \sim \overleftarrow{p}_\varphi(\cdot \mid x_1)$. As the
model trains and begins to better approximate $p_1$, the buffer
becomes populated with samples $x_1$ that are probable under $p_1$.
Thus, the buffer helps the sampler focus on the relevant regions of
the space and retain information about previously
discovered modes.

\paragraph{Reverse trajectories.}
To obtain the batch of trajectories $\tau$ starting at $x_0$, we use
both the reverse trajectory used to produce $x_0$ (see above) and a
batch of $N-1$ trajectories drawn on-policy from
$\overrightarrow{p}_\theta(\tau\mid x_0)$ to form a batch of $N$
trajectories sharing their initial point. The reuse of the backward
trajectories allows the algorithm to learn on trajectories that reach
high-density regions of $p_1$. However, using \textit{only} reverse
trajectories prevents the model from sufficiently exploring the whole
space. Therefore, careful tuning is needed to strike a perfect
balance between exploration and exploitation. We use $N=2$ for all
our experiments.

\paragraph{Langevin updates.} Following the method of
\citet{sendera2024improved} for diffusion samplers, we periodically
update the buffer using a few steps of unadjusted
Langevin on the density $p_1$ to correct for the sampler's imperfect
fit to the target.

\paragraph{Mixing on-policy and off-policy training.} In the training
policy, we use a mixture of initial points $x_0$ and trajectories
$\tau$ sampled on-policy and those sampled using the Langevin-updated
buffer with reverse trajectory reuse, as described above. The
frequency of using the buffer is called the \emph{off-policy ratio},
and we ablate different off-policy ratios in
\Cref{tab:off_policy_ablation}. For most of the experiments we use
constant off-policy ratio 0.8.

We provide details of all off-policy methods in \Cref{appendix:experiments},
and these methods are ablated in \Cref{sec:off-policy-ablation}.

\subsection{Iterative proportional fitting with data-to-energy steps}

The full IPF algorithm for data-to-energy SB alternates between
backward steps that train $\varphi$ to convergence using the
maximum-likelihood objective \eqref{eq:tlm-bwd} and forward steps
that train $\theta$ to convergence using \eqref{eq:d2e-loss}. The
backward step is trained using samples from $p_0$ and forward
trajectories from $\overrightarrow{p}_\theta(\tau\mid x_0)$, while
the forward step is trained using trajectories obtained using the
off-policy methods described above. The complete algorithm is
summarised in \Cref{alg:d2e-ipf}. We reuse the model weights and
preserve the buffer state from the previous IPF step, although we
randomly reinitialise a fraction of samples stored in buffer for the
outsourced experiments.

\paragraph{Energy-to-energy generalisation.} The data-to-energy IPF
algorithm can easily be generalised to the case where samples from
neither $p_0$ nor $p_1$ are available, but both are given by
unnormalised densities $p_i(x)=e^{-\mathcal{E}_i(x)}/Z_i$. In this
case, both the backward and forward IPF steps must be performed using
the variance-based loss \eqref{eq:vargrad}, with appropriate choices
of training distributions (such as keeping separate replay buffers
for both marginals). We call this \emph{energy-to-energy} SB and show
preliminary results in \Cref{sec:exp_d2e}.

\paragraph{Evaluation metrics.}
We consider three metrics for evaluating the approximate solutions to
the SB problem yielded by our data-to-energy IPF algorithm. Because
SB is a constrained optimisation problem, it is necessary to measure
both the constraint satisfaction (\ie, that the solution is a
transport from $p_0$ to $p_1$) and the cost (divergence from the
reference process). To this end we measure ELBO, path KL, and
Wasserstein distance to oracle samples from the target distribution
(when available); see \Cref{appendix:metrics} for details.

\section{Outsourced sampling with Schrödinger bridges}

We describe how our algorithm for solving data-to-energy
Schr\"odinger bridges can be applied to the problem of Bayesian
posterior sampling under a
pretrained generative model prior $p(x)$ by pulling the sampling
problem back to its latent space.

Consider a posterior of the form $p(x \mid y) \propto p(x)r(x, y)$,
where $p(x)$ is a prior over the data space (\eg, images) and $r(x,
y)$ is a constraint function that encodes the conditional information
about the sample $x$ (\eg, a
class likelihood or match to a text prompt).
If the pretrained generative model is expressed as a deterministic
function $f$ of a random noise variable
$z \sim p(z)$, \citet{venkatraman2025outsourced} proposed to sample
the posterior pulled back to the noise space,
with density $p(z\mid y)\propto p(z)r(f(z),y)$, using a diffusion
sampler. If $z$ is distributed with this density,
then samples $f(z)$ follow the desired posterior distribution
$p(x\mid y)$ in data space. Such a method was successfully applied in
the latent spaces of various models types, such as GANs, continuous
normalising flows.

Instead of using a diffusion sampler, we propose to model a
Schrödinger bridge between the distributions $p(z)$ and $p(z \mid
y)$. Since neither the normalising constant nor samples from the
latter distribution are available, we use the data-to-energy
algorithm described in \Cref{sec:d2e_theory}. Modelling a Schrödinger
bridge instead of simply a diffusion sampler has the advantage of
transporting prior samples to nearby posterior samples in latent
space, which is expected to preserve semantic content that is not
constrained by $y$; we show this empirically in
\Cref{sec:outsourced-experiments}.

\paragraph{Metrics for outsourced stochastic transport.}
In order to evaluate the performance of our method on the stochastic
optimal transport task in the latent space, we compute path KL in the
latent space, as well as the $L^2$ static transport cost between
prior samples from $p_0(x_0)$ and the pushed-forward samples from the
approximated posterior $\overrightarrow{p}(\tau \mid x_0) p_0(x_0)$.
For image tasks with a classifier reward, in order to evaluate the
quality of generated images with respect to the target posterior, we
use the mean log-constraint value and FID \citep{heusel2017gans}. The
latter is computed between images decoded from latents sampled from
the trained SB model and images of the target class(es), which are
not available to the model during training.
\begin{table}[t]
  \vspace*{-1em}
  \caption{Comparison of data-to-data IPF methods. \textbf{Bold}
  indicates the best-performing method.}
  \resizebox{\linewidth}{!}{
    \begin{tabular}{@{}lllllll}
      \toprule
      Distributions $\rightarrow$ &  \multicolumn{2}{c}{Gauss
      $\leftrightarrow$ GMM} & \multicolumn{2}{c}{Gauss
      $\leftrightarrow$ Two Moons} & \multicolumn{2}{c}{Two Moons
      $\leftrightarrow$ GMM} \\
      \cmidrule(lr){2-3}\cmidrule(lr){4-5}\cmidrule(lr){6-7}
      Algorithm\ $\downarrow$ Metric $\rightarrow$
      & \multicolumn1c{$\mathcal{W}_2^2 (\downarrow)$} &
      \multicolumn1c{Path KL $(\downarrow)$}
      & \multicolumn1c{$\mathcal{W}_2^2 (\downarrow)$} &
      \multicolumn1c{Path KL $(\downarrow)$}
      & \multicolumn1c{$\mathcal{W}_2^2 (\downarrow)$} &
      \multicolumn1c{Path KL $(\downarrow)$} \\
      \midrule
      DSBM-IMF \citep{shi2023dsbm}                     &
      0.046\std{0.004}            & 1.004\std{0.137}          &
      0.061\std{0.034}            & \textbf{1.813\std{0.960}}    &
      0.043\std{0.012}            & \textbf{1.893\std{0.369}}   \\
      DSBM-IMF+ \citep{shi2023dsbm}                    &
      0.060\std{0.049}            & \textbf{0.909\std{0.426}} &
      0.049\std{0.007}            & 1.848\std{0.285}             &
      0.038\std{0.010}            & \textbf{1.894\std{0.324}}   \\
      ${\rm [SF]^2M}$ \citep{tong2024simulation}       &
      0.041\std{0.021}            & 2.295\std{0.513}          &
      0.053\std{0.022}            & 3.321\std{1.862}             &
      0.057\std{0.030}            & 3.823\std{0.314}            \\
      \midrule
      \textit{IPF-based}\\
      DSB mean \citep{debortoli2021diffusion}          &
      0.093\std{0.096}            & 5.886\std{4.460}          &
      0.111\std{0.041}            & 6.398\std{1.949}             &
      0.078\std{0.029}            & 5.520\std{3.163}            \\
      DSB score \citep{debortoli2021diffusion}         &
      0.052\std{0.018}            & 5.645\std{3.474}          &
      0.171\std{0.149}            & 14.346\std{8.776}            &
      0.066\std{0.035}            & 5.231\std{2.802}            \\
      SDE \citep{chen2021likelihood}                   &
      \textbf{0.037\std{0.010}}   & 2.088\std{1.228}          &
      0.033\std{0.004}            & 2.262\std{0.268}             &
      \textbf{0.025\std{0.010}}   & 3.915\std{0.285}            \\
      LL fixed var. $\approx$ \citet{vargas2021solving}&
      \textbf{0.037\std{0.014}}   & 2.507\std{0.366}          &
      0.033\std{0.005}            & 2.351\std{0.149}             &
      0.031\std{0.011}            & 3.710\std{0.332}            \\
      LL learnt var. (ours)                            &
      \textbf{0.042\std{0.018}}   & 2.840\std{0.668}          &
      \textbf{0.022\std{0.009}}   & 4.288\std{1.876}             &
      \textbf{0.023\std{0.013}}   & 3.938\std{0.526}            \\
      \bottomrule
    \end{tabular}
  }
  \label{tab:comparison_d2d}
  \vspace*{-0.5cm}
\end{table}

\begin{wrapfigure}[12]{r}{0.5\textwidth}
  \centering
  \vspace{-1.3em}
  \includegraphics[height=0.24\textwidth]{
    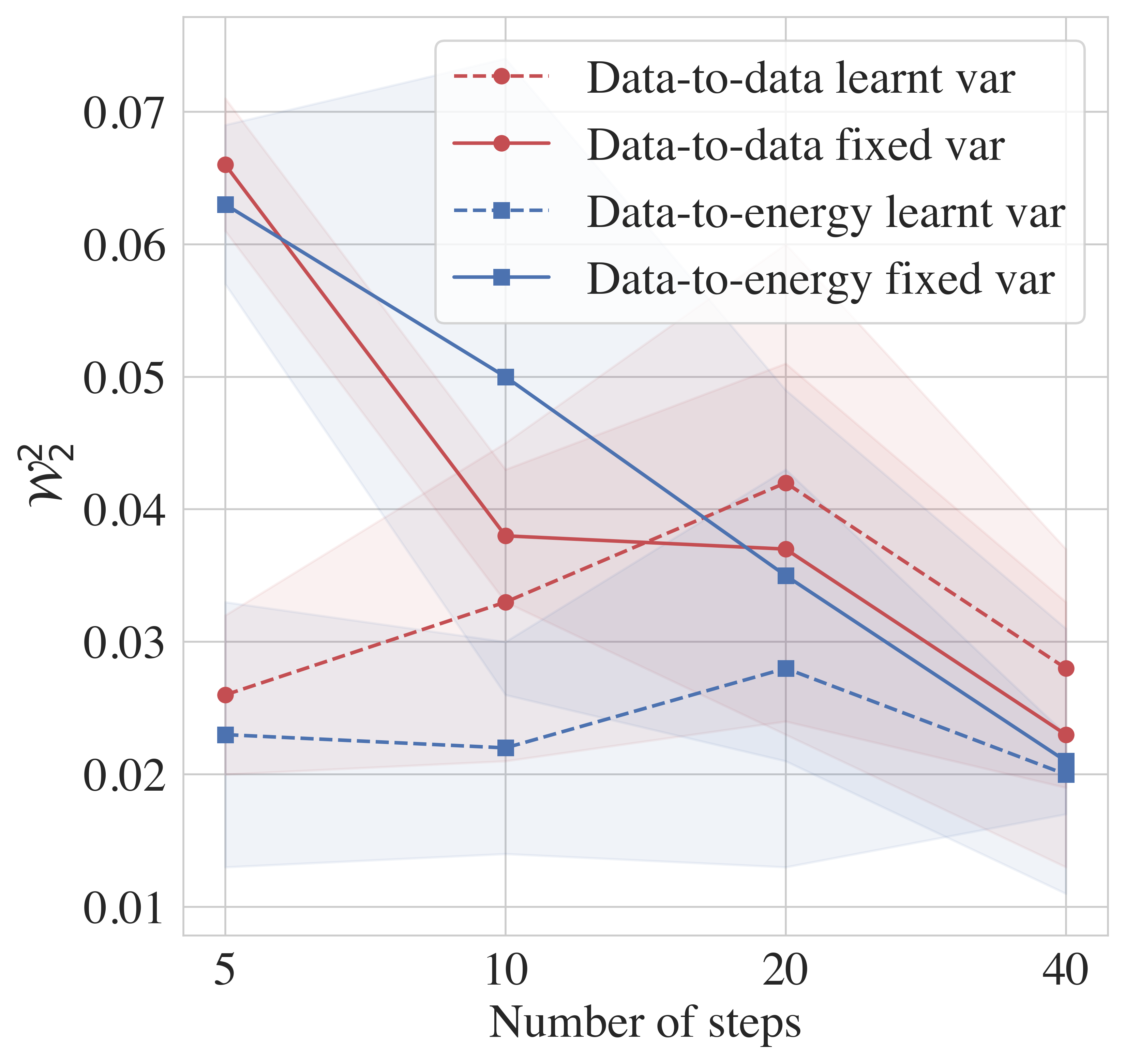
  }
  \includegraphics[height=0.24\textwidth]{
    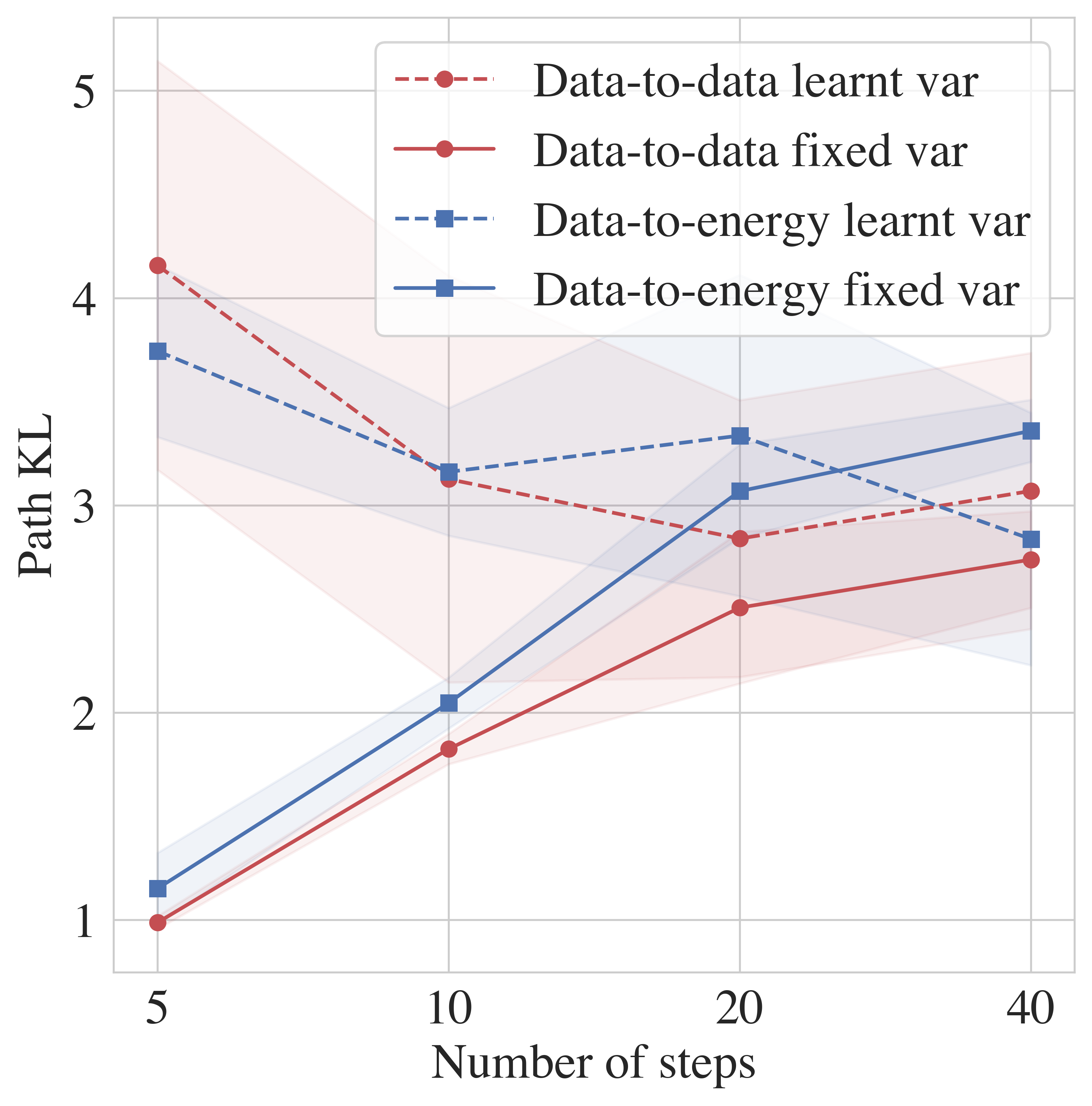
  }
  \caption{
    $\mathcal{W}_2^2$ and Path KL depending on the number of
    discretization steps for
    Gauss $\leftrightarrow$ GMM.
  }
  \label{fig:num_steps_ablation}
\end{wrapfigure}

\section{Experiments}
\subsection{Benefits of trainable variance}
In order to show the benefits of trainable variance in
time-discretised processes, we present the comparisons of existing
data-to-data SB methods with our proposed algorithms, including both
learnt-variance and fixed-variance alternatives. The data-to-data
experiments are conducted on several synthetic 2-dimensional
benchmarks (following \citet{shi2023dsbm}):  Gauss $\leftrightarrow$
GMM, Gauss $\leftrightarrow$ Two Moons, Two Moons $\leftrightarrow$
GMM (where Gauss is an isotropic Gaussian and GMM is a mixture of
eight Gaussians). We compare with \citet{shi2023dsbm} (DSBM and
DSBM++), \citet{debortoli2021diffusion} (DSB), which also uses IPF
for training, \citet{chen2021likelihood} (SDE), which uses a
continuous-time version of IPF, and \citet{tong2024simulation}
(${[\rm SF]^2M}$), which relies on a minibatch approximation to
entropic optimal transport. All experiments use $K=20$ discretisation
steps. The results in \Cref{tab:comparison_d2d} clearly show the
benefits of training the variances, despite the discrete-time
processes not being consistent with an underlying continuous-time process.

We further investigate the effect of learnt variance at varying
numbers of discretisation steps. We find that learnt variance allows
for more accurate modelling in both data-to-energy and data-to-data
settings when the number of steps is small. Results are shown in
\Cref{fig:num_steps_ablation} (in numerical form in
\Cref{tab:num_steps_ablation}).

For all 2-dimensional experiments we use ${\rm d}X_t = \sqrt{2}\,{\rm
d}W_t$ as the reference process; training is done using 4000 steps
for both backward and forward processes and 20 IPF steps.  We use the
same neural network architecture for all 2-dimensional experiments.
We provide a detailed experiment configuration in \Cref{appendix:experiments}.

\subsection{Data-to-energy and energy-to-energy Schrödinger bridge}
\label{sec:exp_d2e}
\begin{wrapfigure}[32]{r}{0.48\textwidth}
  \vspace{-2.25em}
  \begin{tabular}{@{}r@{}c@{}}
    & \makebox[\linewidth]{
      \scriptsize $t = 0$\hspace{0.3cm}
      \raisebox{0.1ex}{\scriptsize$\overset{\overrightarrow{\mathbb{P}}_t}{\xrightarrow{\hspace{4.6cm}}}$}
      \hspace{0.4cm}$t = 1$
    } \\
    \raisebox{0.5cm}{\scriptsize \rotatebox[origin=c]{90}{IPF step 1}}
    & \includegraphics[width=1.0\linewidth,
    height=0.1451\linewidth]{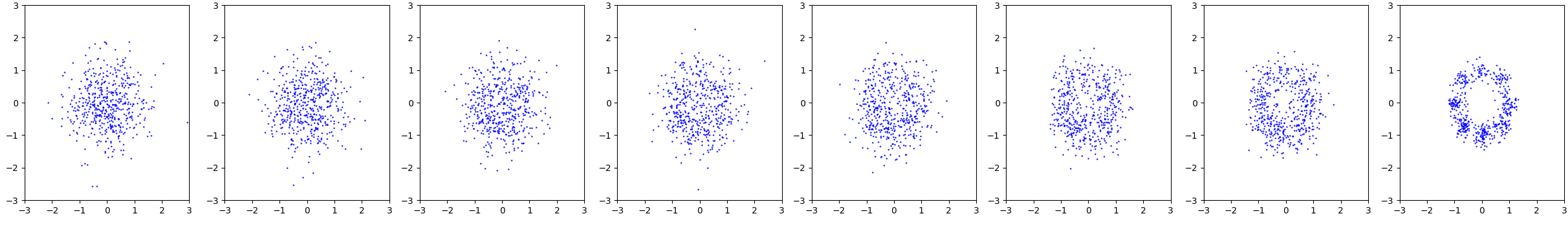} \\
    \raisebox{0.5cm}{\scriptsize \rotatebox[origin=c]{90}{IPF step 7}}
    & \includegraphics[width=1.0\linewidth,
    height=0.1451\linewidth]{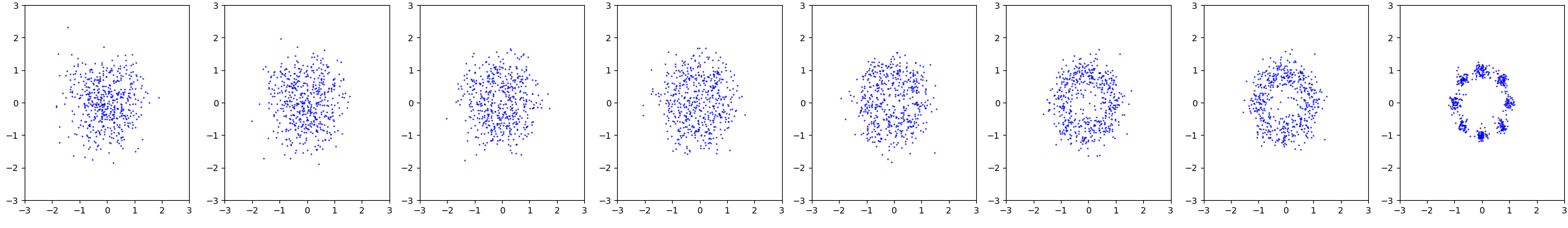} \\
    \raisebox{0.5cm}{\scriptsize \rotatebox[origin=c]{90}{IPF step 20}}
    & \includegraphics[width=1.0\linewidth,
    height=0.1451\linewidth]{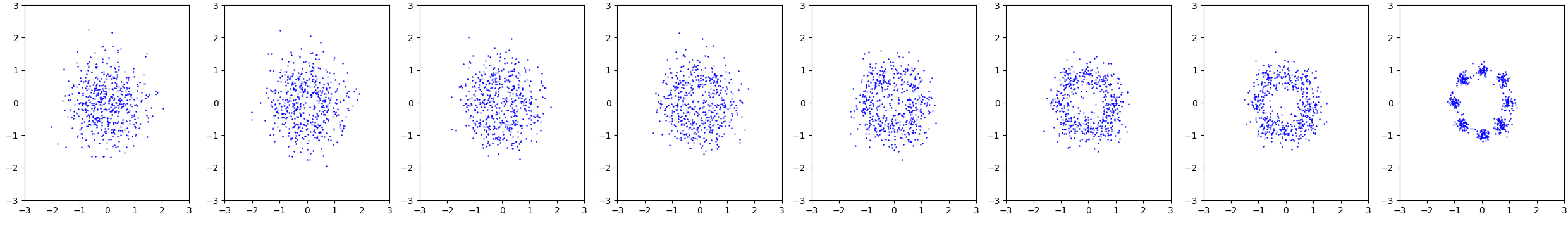}
  \end{tabular}

  \hrule
  \vspace{0.125cm}

  \begin{tabular}{@{}r@{}c@{}}
    \raisebox{0.5cm}{\scriptsize \rotatebox[origin=c]{90}{IPF step 1}}
    & \includegraphics[width=1.0\linewidth,
    height=0.1451\linewidth]{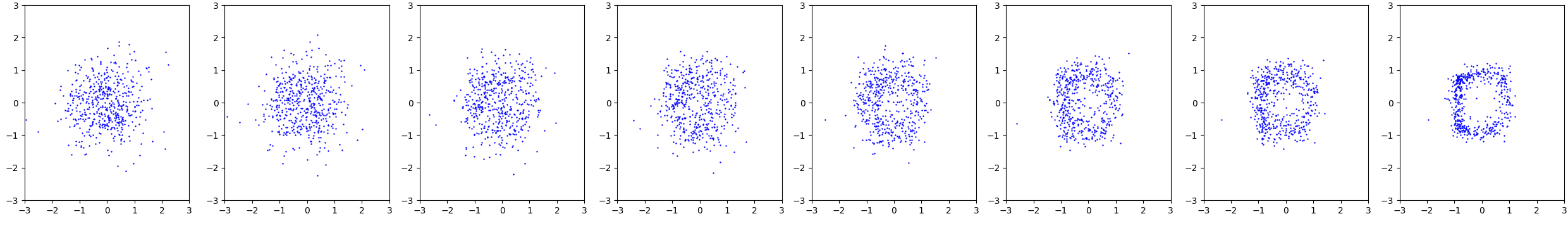} \\
    \raisebox{0.5cm}{\scriptsize \rotatebox[origin=c]{90}{IPF step 7}}
    & \includegraphics[width=1.0\linewidth,
    height=0.1451\linewidth]{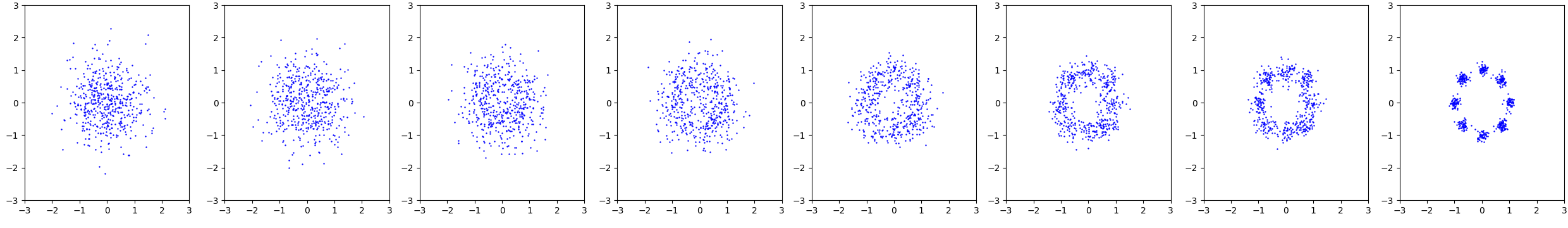} \\
    \raisebox{0.5cm}{\scriptsize \rotatebox[origin=c]{90}{IPF step 20}}
    & \includegraphics[width=1.0\linewidth,
    height=0.1451\linewidth]{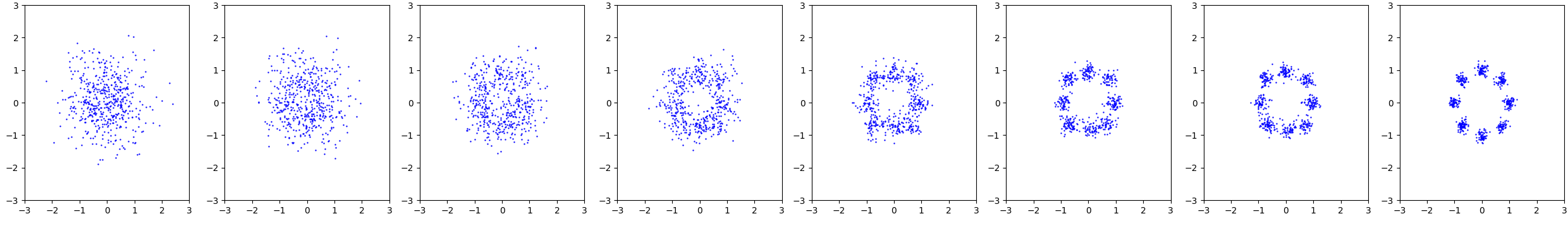}
  \end{tabular}

  \hrule
  \vspace{0.125cm}

  \begin{tabular}{@{}r@{}c@{}}
    \raisebox{0.5cm}{\scriptsize \rotatebox[origin=c]{90}{IPF step 1}}
    & \includegraphics[width=1.0\linewidth,
    height=0.1451\linewidth]{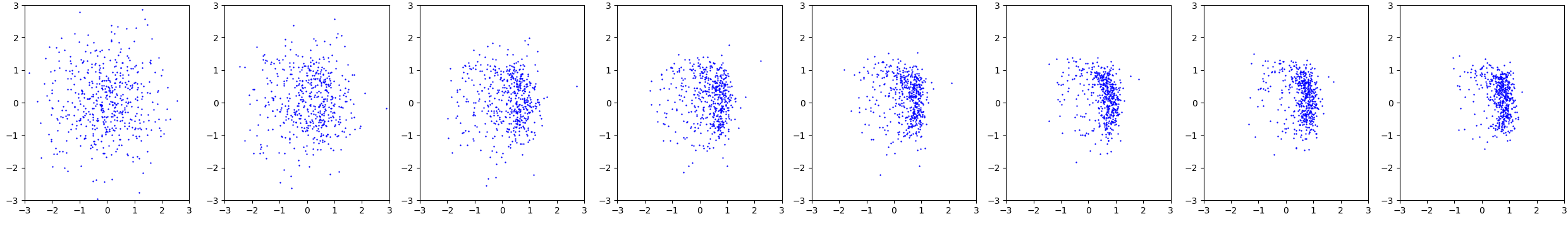} \\
    \raisebox{0.5cm}{\scriptsize \rotatebox[origin=c]{90}{IPF step 7}}
    & \includegraphics[width=1.0\linewidth,
    height=0.1451\linewidth]{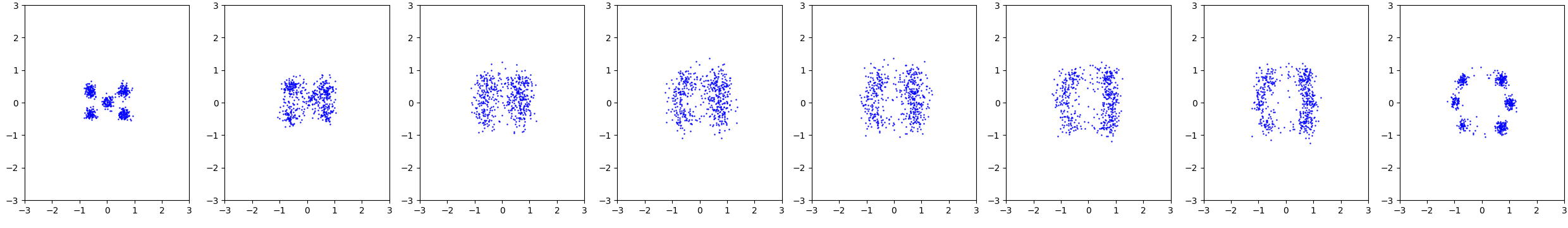} \\
    \raisebox{0.5cm}{\scriptsize \rotatebox[origin=c]{90}{IPF step 20}}
    & \includegraphics[width=1.0\linewidth,
    height=0.1451\linewidth]{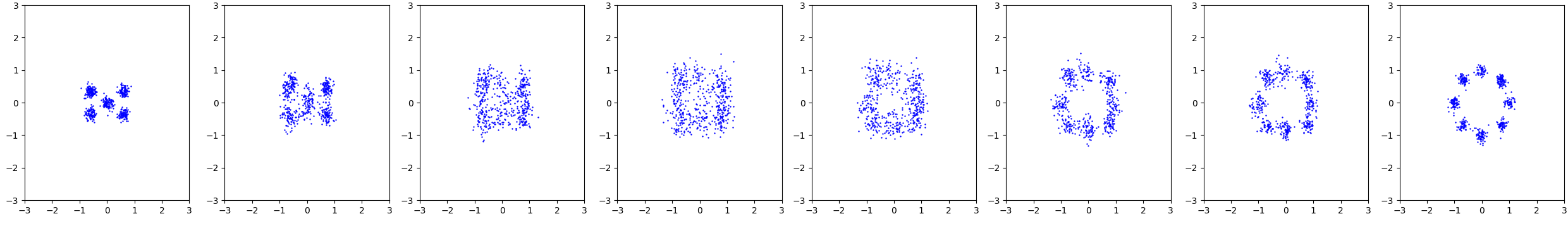}
  \end{tabular}
  \vspace*{-0.75em}
  \caption{
    Comparison of learnt processes at various IPF iterations for
    data-to-data (\textit{top}), data-to-energy (\textit{middle}) and
    energy-to-energy (\textit{bottom}) settings. For the
    energy-to-energy setting we use two different mixtures of
    Gaussian distributions.
  }
  \vspace*{-2.0cm}
  \label{fig:d2e}
\end{wrapfigure}
To prove the viability of data-to-energy Schrödinger bridge training
we compare the bridge between Gaussian and GMM distributions trained
using data-to-data and data-to-energy IPF versions (where the
  data-to-energy version has no access to the data samples that the
data-to-data algorithm sees). \Cref{fig:d2e} shows the resulting IPF
trajectories and \Cref{tab:num_steps_ablation} shows that the
data-to-energy model is comparable to one trained with data samples.

\subsection{Outsourced Schrödinger bridge}
\label{sec:outsourced-experiments}

\looseness=-1
Finally, we show the scalability of our method by running
data-to-energy Schrödinger bridge algorithm in the latent space of
generative model. We use the StyleGAN \citep{karras2020analyzing,
karras2021alias} and SN-GAN \citep{miyato2018spectral} generators
trained on CIFAR-10 \citep{Krizhevsky09learningmultiple}, and VAE
\citep{kingma2013auto, rezende2014stochastic} generator trained on
MNIST \citep{lecun1998gradient}. We train the bridge model between
the latent space prior $p(z)$, which in our case is always a Gaussian
distribution, and a reward-reweighted prior of the form $r(f(z), y)
\cdot p(z)$. The reward function $r(x, y)$ is a classifier that
returns the probability of the object $x$ belonging to the class $y$
(\ie, the probability that $x$ is, for example, a boat for CIFAR-10
or the digit 5 for MNIST).

MNIST experiments are conducted in two setups: (a) the reward function
returns the probability that $x$ is even or odd (b) the reward function
returns the probability that $x=5$.
For CIFAR-10 experiments we use StyleGAN \citep{karras2020analyzing,
karras2021alias} generator with latent dimension 512 and SN-GAN
\citep{miyato2018spectral} generator with latent dimension 128. We
use a pretrained classifier model as a reward function. Samples are
shown in \Cref{fig:osb}; more CIFAR-10 results can be found in
\Cref{appendix:visual}.

In \Cref{tab:gan_comparison} we compute FID between samples of the
target class posterior obtained from the trained SB and images
belonging to the target class in the dataset; we use the same
metric for a set of ground truth posterior samples obtained by
rejection sampling. Remarkably, the transported samples tend to have
lower FID than samples from the true distribution.

These results reveal a benefit of training a Schrödinger bridge
model, as opposed to a diffusion sampler or an arbitrary stochastic
mapping. It can be seen in \Cref{fig:osb} that images already
belonging to the target class change little, while those belonging to
other classes maintain features that are unrelated to the target
class: the background and global structure are preserved. This
suggests that style transfer for higher-dimensional images can be a
promising application of our method.
\begin{figure}[t]
  \vspace*{-1em}
  \centering
  \label{fig:gan_grid_comparison}
  \begin{tabular}{@{} *{12}{c@{}}}
    \toprule
    \multicolumn{8}{c}{\bf SN-GAN} & \multicolumn{4}{c}{\bf StyleGAN} \\
    \cmidrule(lr){1-8}\cmidrule(lr){9-12}
    \multicolumn{2}{c}{Car} & \multicolumn{2}{c}{Cat} &
    \multicolumn{2}{c}{Dog} &
    \multicolumn{2}{c}{Horse} & \multicolumn{2}{c}{Horse} &
    \multicolumn{2}{c}{Truck} \\
    \cmidrule(lr){1-2}\cmidrule(lr){3-4}\cmidrule(lr){5-6}\cmidrule(lr){7-8}\cmidrule(lr){9-10}\cmidrule(lr){11-12}
    {\small Prior} & {\small Posterior} & {\small Prior} & {\small
    Posterior} & {\small Prior} & {\small Posterior} &
    {\small Prior} & {\small Posterior} & {\small Prior} & {\small
    Posterior} & {\small Prior} & {\small Posterior} \\
    \includegraphics[width=0.08\linewidth, trim = 0 0 0 0, clip]{
    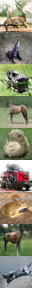} &
    \includegraphics[width=0.08\linewidth, trim = 0 0 0 0, clip]{
    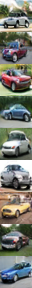} &
    \includegraphics[width=0.08\linewidth, trim = 0 0 0 0, clip]{
    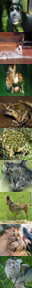} &
    \includegraphics[width=0.08\linewidth, trim = 0 0 0 0, clip]{
    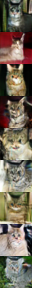} &
    \includegraphics[width=0.08\linewidth, trim = 0 0 0 0, clip]{
    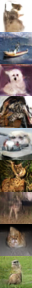} &
    \includegraphics[width=0.08\linewidth, trim = 0 0 0 0, clip]{
    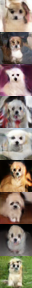} &
    \includegraphics[width=0.08\linewidth, trim = 0 0 0 0, clip]{
    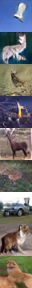} &
    \includegraphics[width=0.08\linewidth, trim = 0 0 0 0, clip]{
    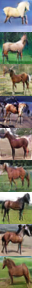} &
    \includegraphics[width=0.08\linewidth, trim = 0 0 0 0, clip]{
    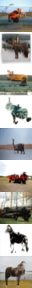} &
    \includegraphics[width=0.08\linewidth, trim = 0 0 0 0, clip]{
    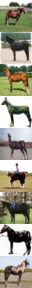} &
    \includegraphics[width=0.08\linewidth, trim = 0 0 0 0, clip]{
    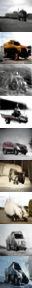} &
    \includegraphics[width=0.08\linewidth, trim = 0 0 0 0, clip]{
    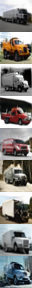} \\
    \bottomrule
  \end{tabular}
  {
    \centering
    \begin{minipage}[t]{0.32\textwidth}
      \begin{tabular}{@{}c@{\hspace{0.05cm}}c@{}}
        Prior & Fives \\
        \includegraphics[width=2.2cm]{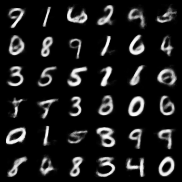}
        &
        \includegraphics[width=2.2cm]{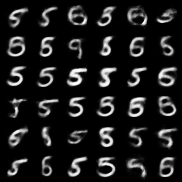}
        \\
      \end{tabular}
    \end{minipage}
    \hspace{0.05cm}
    \begin{minipage}[t]{0.32\textwidth}
      \begin{tabular}{@{}c@{\hspace{0.05cm}}c@{}}
        Prior & Even \\
        \includegraphics[width=2.2cm]{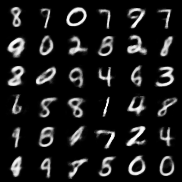}
        &
        \includegraphics[width=2.2cm]{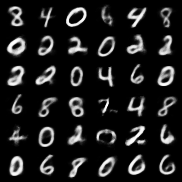}
        \\
      \end{tabular}
    \end{minipage}
    \hspace{0.05cm}
    \begin{minipage}[t]{0.32\textwidth}
      \begin{tabular}{@{}c@{\hspace{0.05cm}}c@{}}
        Prior & Odd \\
        \includegraphics[width=2.2cm]{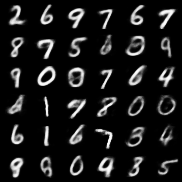}
        &
        \includegraphics[width=2.2cm]{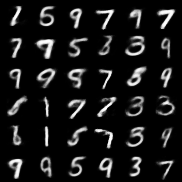}
        \\
      \end{tabular}
    \end{minipage}
  }
  \vspace*{-1em}
  \caption{
    Outsourced Schrödinger bridge on MNIST and CIFAR-10. The bridge
    preserves style features (thickness, background colour,
    orientation) while transforming digits to the target class.\label{fig:osb}
  }
  \vspace*{-1em}
\end{figure}
\begin{table}[!t]
  \centering
  \begin{minipage}{0.29\linewidth}
    \caption{
      \looseness=-1
      ELBO ($\uparrow$) values and FID ($\downarrow$) scores
      comparing CIFAR-10 samples from the posterior under a GAN prior
      and a classifier to data samples of the target class.
      In this setting diffusion sampler performed worse than other
      methods, therefore,
      FID scores are computed only for `Dog' and `Car' classes.
    }
    \label{tab:gan_comparison}
  \end{minipage}\hfill
  \begin{minipage}{0.70\linewidth}
    \vspace{-1.3em}
    \resizebox{\linewidth}{!}{
      \begin{tabular}{@{}lcccccc@{}}
        \toprule
        & \multicolumn{4}{c}{SN-GAN} & \multicolumn{2}{c}{StyleGAN} \\
        \cmidrule(lr){2-5}\cmidrule(lr){6-7}
        Algorithm $\downarrow$ Class $\rightarrow$ & Car & Cat & Dog &
        Horse & Horse & Truck \\
        \midrule
        Same class          & 10.371 &  17.950 & 15.034 & 12.876 &
        12.876 & 9.289  \\
        \midrule
        Rejection sampling  & 31.334 & 42.242 & 43.691 & 35.019 &
        97.855  & 76.403 \\
        Langevin            & 25.296 & 33.619 & 37.665 & 27.601 &
        85.611  & 68.678 \\
        Diffusion sampler   & 83.940 & - & 60.512 & - & - & - \\
        Outsourced SB       & 22.312 & 40.489 & 37.287 & 33.021 &
        58.988  & 55.346 \\

        \midrule[1.0pt]
        & \multicolumn{6}{c}{ELBO}  \\
        \midrule

        Outsourced SB       & -6.737 & -9.356 & -6.087 & -5.961  &
        -21.784 & -6.809 \\
        \bottomrule
      \end{tabular}
    }
  \end{minipage}
\end{table}

\clearpage\newpage
\subsection{Ablations of off-policy techniques}
In order to verify our design choices for the high-dimensional
experiments we provide a detailed ablation of the various off-policy
reinforcement learning tricks described in \Cref{sec:off-policy}. The
results are given in \Cref{tab:off_policy_ablation}. We use a
simple on-policy setup as a baseline.
We find that saving samples in a replay buffer (\textit{buffer}) to
use them for sampling backward trajectories and applying Langevin
update to the buffer samples (\textit{buffer + Langevin})
significantly improves the Path KL metrics, therefore yielding a
bridge closer to the optimal one.
Moreover, we show that reusing backward trajectories for the
computation of loss also improves Path KL, giving the best model
based on this metric.
However, reusing backward trajectories negatively impacts the mean
log-reward metric, possibly because it prevents mode collapse. To
balance between modelling modes and achieving low transport cost, we
explore the possibilities of both setting a smaller off-policy ratio
-- fraction of off-policy trajectories -- and annealing the
off-policy ratio throughout the training; the latter sometimes
produces improvements. All ablation experiments are conducted using
the SN-GAN generator and VGG13 classifier on the CIFAR-10 dataset. All
models are trained to amortise sampling from one class (Dogs) and are
trained with the same seed.

\label{sec:off-policy-ablation}
\begin{table}[t]
  \caption{
    Ablation of off-policy reinforcement learning techniques on the
    SN-GAN outsourced sampling problem.
    \textbf{Bold} indicates the best result, \underline{underlined}
    indicates second best. The model is trained to amortise sampling
    from one class (dogs).
  }
  \resizebox{\linewidth}{!}{
    \begin{tabular}{@{}lcccc}
      \toprule
      Algorithm\ $\downarrow$ Metric $\rightarrow$ & ELBO
      $(\uparrow)$ & Path KL $(\downarrow)$ & $L^2_2(x_0, x_1)
      (\downarrow)$  & mean log-reward $(\uparrow)$\\

      \midrule
      on-policy                       & $ -190.920 $               &
      1506.407              & 10.949                & $ -0.233 $             \\
      \midrule
      buffer         & \underline{$ -188.351 $}   & 622.895
      & 10.957                & $ \bf-0.125 $    \\
      \quad + Langevin          & $ -188.554 $               &
      383.514               & \underline{10.130}    & $ -0.286 $             \\
      \quad\quad + reuse backward trajectory  & {$ \bf-188.149 $}
      & \textbf{206.094}       & \textbf{10.046}       & $ -0.657 $
      \\
      \quad\quad\quad + annealed off-policy ratio      & $ -188.355 $
      & \underline{244.270}   & 11.386                & $ -0.149 $
      \\
      \quad\quad\quad smaller off-policy ratio     & $ -188.620 $
      & 668.255               & 11.027                & \underline{$-0.131$} \\
      \bottomrule
    \end{tabular}
  }
  \label{tab:off_policy_ablation}
\end{table}

\subsection{Comparison with analytical solution}
\begin{wrapfigure}[6]{r}{0.45\textwidth}
  \centering
  \vspace{-4.5em}
  \includegraphics[width=1.0\linewidth,trim=100 0 100 40,clip]{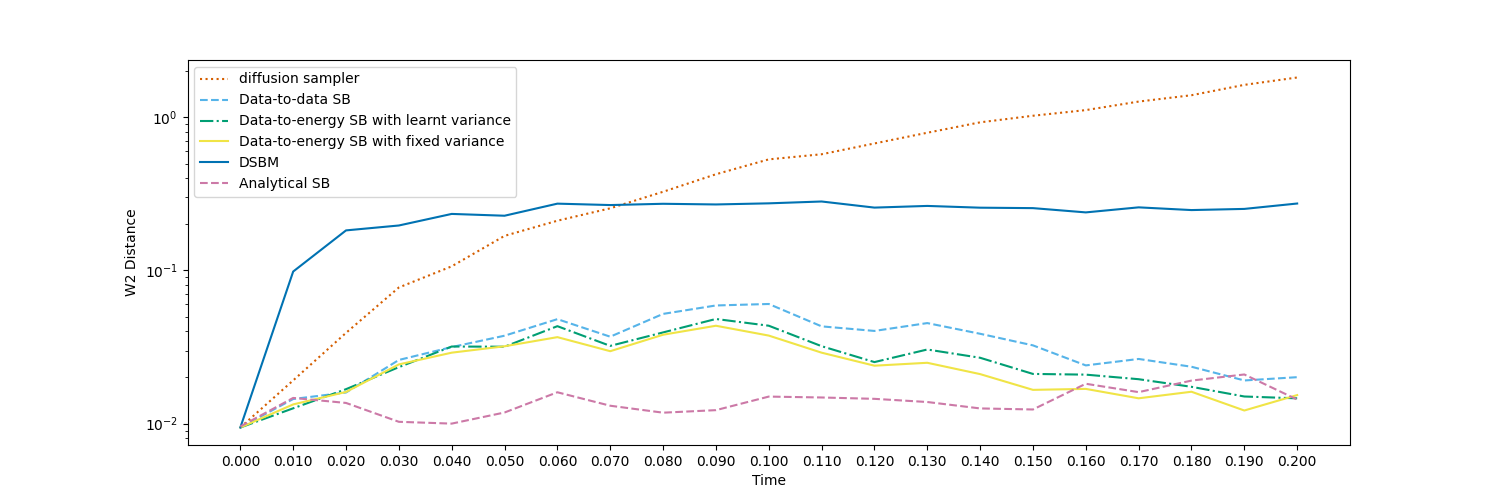}\\[-0.5em]
  \caption{SB algorithms' comparison in $\mathcal{W}_2$ to the analytical solution.}
  \label{fig:sb-analytic}
\end{wrapfigure}
\looseness=-1
To validate the proposed SB algorithm, we apply it to the problem of
finding the SB between two Gaussians. In this case, the problem has
an analytical solution \citep{bunne2023schrodinger}. For our
experiments, we train SB between $\mathcal{N}\left( 0, I\right)$  and
$\mathcal{N}\left(
  \left[ \begin{smallmatrix} 1 \\ 1 \end{smallmatrix} \right],
  \left[ \begin{smallmatrix} 3.0 & 1.2 \\ 1.2 & 1.0 \end{smallmatrix} \right]
\right)$.
We compare the data-to-energy and data-to-data SB algorithms described in
this paper, as well as DSBM \citep{debortoli2021diffusion}, to the SB
computed analytically. For comparison, we use the Wasserstein-2
distance computed between samples obtained from the SB algorithm and
analytical solution, for each time step of the trajectory. The
results are in \Cref{fig:sb-analytic}.

\section{Conclusion}
This paper shows the potential of training data-to-energy Schrödinger
bridges in a time discretisation with learnable drift and variance
for the forward and backward processes. We showed that, despite its
complexity, our method can be successfully scaled. Future work should
focus more on scaling the data-to-energy Schrödinger bridges to
higher dimensions, as well as arbitrary prior distributions, which
would significantly improve the versatility of the proposed
algorithms. Moreover, the trained samplers are prone to mode
collapse, therefore, future work should investigate techniques to
further improve mode coverage. One more promising direction would be
to amortise over the distribution of conditions in image generation
problems, instead of learning a model for each specific condition.
Finally, we are excited to explore various domains in which the
proposed algorithms can be applied. Interesting areas include
text-conditional image reward fine-tuning of diffusion models and
discrete Schrödinger bridge problems (recently explored in \citet{carter2026discrete}).

\newpage

\section*{Acknowledgements}

The authors thank Sanghyeok Choi for comments on the manuscript.

This work was enabled by the computational resources of the
Edinburgh International Data Facility (EIDF) with funding from the
Edinburgh Generative AI Laboratory (GAIL).

ESW acknowledges support from the CIFAR Learning in Machines and
Brains programme.

\newpage
\appendix
\section{Related works}\label{appendix:related_work}
In this section we establish links between our method and other
research directions in the literature.

\paragraph{Optimal transport.}
Optimal transport is a well-established area of research with a solid
theoretical background and scalable applied algorithms. The problem
is concerned with finding the optimal transportation map which
minimises a given transportation cost. Originally, the problem was
proposed by Monge \citep{monge1781memoire}. In 20th century this
problem was reformulated and generalised by Kantorovich in a series
of works including \citet{kantorovich1958space,
kantorovich1960mathematical}. Since then the problem has been
rigorously studied, refer to \citet{peyre2019computational,
villani2008optimal} for a detailed presentation of theory. In a
discrete case optimal transport can be solved using Sinkhorn's
algorithm \citep{sinkhorn1964relationship, peyre2019computational}.
Recent works have applied optimal transport to a series of machine
learning problems, including image-to-image translation
\citep{korotin2022neural}, voice conversion
\citep{asadulaev2024optimal}, and super-resolution \citep{gazdieva2025optimal}.

\paragraph{Schrödinger bridges.} The Schrödinger bridge problem is
concerned with finding stochastic optimal transport dynamics between
two distributions. The problem was originally proposed by Schrödinger
\citep{schrodinger1931umkehrung, schrodinger1932theorie}. The
Schrödinger bridge problem can be seen as a regularised version of
dynamic optimal transport \citep{leonard2013survey} and has
interesting connections to optimal control theory
\citep{chen2021optimal}. Computationally, the problem can be solved
using Iterative Proportional Fitting (IPF) algorithm
\citep{fortet1940resolution, deming1940least,
sinkhorn1964relationship}. \citet{debortoli2021diffusion,
vargas2021solving} proposed the scalable formulation of this method
that allows to compute Schrödinger bridge between a pair of
distributions given by unbiased samples. \citet{chen2021likelihood}
proposes a continuous-time variant of IPF. Methods distinct from IPF
have also been proposed \citep{shi2023dsbm,tong2024simulation}; all
of them assume access to samples from the target distribution for an
unbiased objective.

\paragraph{Diffusion samplers.} Data-to-energy Schrödinger bridge is
related to the problem of sampling from an unnormalised density.
Diffusion samplers \citep{zhang2022path, vargas2023denoising,
richter2024improved, berner2024optimal, blessing2025underdamped}
represent one of the approaches that solve this problem.
Some methods use off-policy reinforcement learning techniques
\citep{lahlou2023theory, sendera2024improved} to amortise sampling
from intractable density. The theoretical connection among various
objectives was established in \citet{berner2025discrete}.

\paragraph{Outsourced sampling.} The concept of outsourced diffusion
sampling -- modelling continuous-time dynamics in latent space for
posterior inference  under pretrained priors -- was proposed in
\citet{venkatraman2025outsourced}. The work shows that sampling can
be efficiently conducted in the latent space of a generator, where
the density landscape is smoother.

\section{Metrics for data-to-energy SB}
\label{appendix:metrics}

First, \emph{if both samples from $p_0$ and the density of $p_0$ are
available}, we evaluate the quality of the learnt
$\overrightarrow{p}_\theta$ as a sampler of $p_1$ using the evidence
lower bound:
\[
  {\rm ELBO} =  \mathbb{E}_{
    x_0 \sim p_0(x_0), \tau \sim \overrightarrow{p}_\theta(\tau \mid x_0)
  } \left[ \log \frac{\overleftarrow{p}_\varphi(\tau \mid x_1)
    \exp(-\mathcal{E}_1(x_1))}{\overrightarrow{p}_\theta(\tau \mid x_0)
  p_0(x_0)} \right]
  \leq \log Z,
\]
which equals the true $\log Z=\log\int\exp(-\mathcal{E}_1(x))\,{\rm
d}x$ if and only if the processes $\overrightarrow{p}_\theta$ and
$\overleftarrow{p}_\varphi$ coincide.

Second, \emph{if samples from $p_1$ are available} (even if not
available to the learner during training), we report the
2-Wasserstein distance $\mathcal{W}_2$ between batches of true
samples from $p_1(x_1)$ and samples obtained from the learnt model
$p_{\theta}(\tau \mid x_0)$, which measures the discrepancy between
the target and modelled marginals.

Third, to approximate the cost, we compute the path KL in discrete time:
\begin{align}
  {\rm KL}(p_0\otimes \overrightarrow{\mathbb{P}}_{t\mid 0} \,\|\,
  p_0\otimes \mathbb{Q}_{t\mid 0})
  &\approx
  {\rm KL} \left( p_0(x_0)\otimes \overrightarrow{p}_\theta(\tau \mid
  x_0) \,\|\, p_0(x_0)\otimes q(\tau\mid x_0) \right) \\
  &= \mathbb{E}_{x_0\sim p_0,\tau \sim
  \overrightarrow{p}_\theta(\tau\mid x_0)} \left[\log
    \frac{\overrightarrow{p}_\theta(\tau\mid x_0)}{q(\tau\mid x_0)}
  \right]\label{eq:kl_highvar}\\
  &=\mathbb{E}_{x_0\sim p_0,\tau \sim
  \overrightarrow{p}_\theta(\tau\mid x_0)} \left[
    \sum_{k=0}^{K-1}{\rm KL}(\overrightarrow{p}_\theta(x_{k+1}'\mid
    x_k)\,\|\,q(x_{k+1}'\mid x_k))
  \right]\label{eq:kl_lowvar}
\end{align}
where $q(\tau)$ is the time discretisation of the reference process
$\mathbb{Q}_t$. The estimator using transition KLs in
\eqref{eq:kl_lowvar} can be seen to be a Rao-Blackwellised
(lower-variance) variant of the estimator in \eqref{eq:kl_highvar},
and we use it because the KL can be computed analytically (as all
transition kernels are Gaussian). It can be shown
(\Cref{appendix:kl}) that this path KL is equivalent to path energy
as used in \citet{shi2023dsbm}.

\section{Relation between path KL and path energy}\label{appendix:kl}
In this section we explain the relation between path KL and path
energy \citep{shi2023dsbm}. Assuming that the transition kernels are given by:
\begin{align}
  \overrightarrow{p}_\theta(x_{k+1}'\mid x_k) &= \mathcal{N} \left(
  x_{k} + v_{\theta}(x_{k}, k\Delta t)\Delta t, \sigma^2 \Delta t I \right) \\
  q(x_{k+1}'\mid x_k) &= \mathcal{N} \left( x_{k} , \sigma^2 \Delta t I\right)
\end{align}
where $v_{\theta}(x_t, t)$ is a learnt drift, $\sigma^2$ is constant
and is the same for both $\mathbb{P}_{t\mid 0}$ and
$\mathbb{Q}_{t\mid 0}$, time-discrete path KL can be written in the
following form:
\begin{align}
  &{\rm KL} \left( p_0(x_0)\otimes \overrightarrow{p}_\theta(\tau
  \mid x_0) \,\|\, p_0(x_0)\otimes q(\tau\mid x_0) \right) \\
  & = \mathbb{E}_{x_0\sim p_0,\tau \sim
  \overrightarrow{p}_\theta(\tau\mid x_0)} \left[\log
    \frac{\overrightarrow{p}_\theta(\tau\mid x_0)}{q(\tau\mid x_0)}
  \right] = \mathbb{E}_{x_0\sim p_0,\tau \sim
  \overrightarrow{p}_\theta(\tau\mid x_0)} \left[
    \sum_{k=0}^{K - 1}\log
    \frac{\overrightarrow{p}_\theta(x_{k+1}'\mid x_k)}{q(x_{k+1}'\mid x_k)}
  \right] \\
  & = \mathbb{E}_{x_0\sim p_0,\tau \sim
  \overrightarrow{p}_\theta(\tau\mid x_0)} \left[
    \sum_{k=0}^{K - 1} \left(
      \frac{v_{\theta}(x_k, k\Delta t)}{\sigma^2} (x_{k + 1} - x_k)
      - \frac{\| v_{\theta}(x_k, k\Delta t) \| ^ 2}{2 \sigma^2} \Delta t
    \right)
  \right]
\end{align}

Given that $x_{k + 1} - x_k=v_{\theta}(x_{k}', k\Delta t)\Delta t +
\sigma\sqrt{\Delta t} \xi_k$, $\xi_k \sim \mathcal{N}(0, I)$ for $0
\leq k \leq K - 1$
and the $\xi_{k_i}$ are independent, the path KL can be finally written as:
\begin{align}
  &{\rm KL} \left( p_0(x_0)\otimes \overrightarrow{p}_\theta(\tau
  \mid x_0) \,\|\, p_0(x_0)\otimes q(\tau\mid x_0) \right) \\
  & = \mathbb{E}_{x_0\sim p_0,\tau \sim
  \overrightarrow{p}_\theta(\tau\mid x_0)} \left[
    \sum_{k=0}^{K - 1} \frac{\| v_{\theta}(x_k, k\Delta t) \| ^ 2}{2
    \sigma^2} \Delta t
  \right] \xrightarrow[\Delta t \to 0]{}
  \frac{1}{2\sigma^2}\mathbb{E}_{ p_0(x_0)\otimes
  \overrightarrow{p}_\theta(\tau \mid x_0)} \left[
    \int \| v_{\theta}(x_t, t) \| ^ 2 {\rm d}t
  \right]
\end{align}
The limit is justified by the Girsanov theorem
\citep{sarkka2019applied}. This yields the path energy used in
\citet{shi2023dsbm}.

\section{Experiment details}\label{appendix:experiments}

\subsection{Data-to-data experiments}
All data-to-data experiments are conducted under the unified setup.
For the neural network we use an MLP with 3 hidden layers and 64
neurons in each layer; each layer is followed by a
LayerNorm\citep{ba2016layer} and SiLU activation function. All neural
networks are trained using AdamW optimiser with learning rate
$0.0008$. Sampling is done in $20$ steps with $t_{\rm max}=0.2$ and
${\rm d}t = 0.01$. We train forward and backward models for 4000
steps at each IPF iteration and we train each model for 20 IPF
iterations. $\rm [SF]^2M$ is trained using $160,000$ steps. The
metrics are computed using $10,000$ samples from the target
distributions and $10,000$ samples obtained from the learnt forward
process. All the metrics are averaged over 5 seeds (42, 43, 44, 45, 46).

\subsection{2D data-to-energy experiments}
For the data-to-energy experiments we use the same neural networks as
for data-to-data experiments. We use $20$ steps for sampling with
$t_{\rm max}=0.8$ and ${\rm d}t = 0.04$. Neural networks are
optimised with AdamW optimiser with learning rate $0.0005$. When
Langevin update is used, we update buffer samples every $500$ steps
during training of the forward process. Langevin is used with the
step size $0.01$ and we do 50 updates each time. We use $2$
trajectories from each $x_0$ for the computation of the VarGrad loss.
All the metrics are averaged over 5 seeds (42, 43, 44, 45, 46).

\subsection{2D energy-to-energy experiments}
We provide a description of energy-to-energy experiment shown in
\Cref{fig:d2e}. We use GMM with 5 modes for the distribution $p_0$
and GMM with 8 modes for distribution $p_1$. For both distributions
we rely exclusively on the corresponding log-densities. We do not use
samples from either $p_0$ or $p_1$. We keep replay buffers for both
densities. We initialise both replay buffers with Gaussian noise at
the beginning of training. Since samples are unavailable, we use the
objective \Cref{eq:vargrad} to learn the forward process and a
similar objective:
\begin{equation}
  \mathcal{L}_{\rm LV}(x_1,\varphi)
  = {\rm Var} \left(
    \log \frac{\overleftarrow{p}_{\varphi}(\tau\mid
    x_1)}{\overrightarrow{p}_{\theta}(\tau\mid x_0)p_0(x_0)}
  \right)
  ={\rm Var} \left(
    \sum_{k=1}^{K} \log \frac{\overleftarrow{p}_{\varphi}(x_{(k -
    1)\Delta t} \mid x_{k\Delta t} )}
    {\overrightarrow{p}_{\theta}(x_{k\Delta t} \mid x_{(k - 1)\Delta
    t} ) } + \mathcal{E}_0(x_0)
  \right),
\end{equation}
to learn the backward process.

\subsection{Outsourced Schrödinger bridge experiments}
\paragraph{Experiments on MNIST.}
For MNIST experiments, we use a custom VAE with 3 layers in both the
decoder and encoder. We use a custom MNIST classifier as a reward
model, which consists of 3 MLP layers. Each layer is followed by the
ReLU activation function, except for the last one, which uses a
sigmoid. We train the forward and backward networks for $5,000$ steps
during each IPF iteration, with 20 IPF iterations in total. All the
networks are trained with AdamW optimiser using learning rate
$0.0008$. We do not use Langevin updates for this experiment, relying
only on a replay buffer. We use the same MLPs as in 2D data-to-energy
experiment to parameterise the backward and forward drift and variance.

\paragraph{Experiments on CIFAR-10 with SN-GAN and StyleGAN.} For the
CIFAR-10 experiments, we use MLP with 3 hidden layers, which has 256
hidden units for the SN-GAN experiments and 512 for the StyleGAN
experiments. We train forward network for $500$ steps and backward
network for $100$ steps during each IPF iteration, for a total of
$300$ IPF iterations. All the networks are trained with AdamW
optimiser using learning rate $0.0005$. Langevin updates are made
every $500$ iterations during the training of the forward network. We
run Langevin for $500$ steps with initial step size of $0.01$ and
anneal step size to $0.001$ during the updates.

We use $20$ steps for sampling with ${\rm d}t=0.04$ for the StyleGAN
experiments and ${\rm d}t=0.005$ for SN-GAN experiments. We use
Wiener process, ${\rm d}X_t = \sqrt{2}{\rm d}W_t$, as the reference
process. All the main experiments are conducted with a replay buffer
and Langevin updates, with off-policy ratio of 0.8, and the backward
trajectories are reused for computing VarGrad loss.

For the reward model we use VGG \citep{simonyan2015very} classifier
pretrained on CIFAR-10. The weights are taken from {\tt
\url{https://github.com/huyvnphan/PyTorch_CIFAR10}}. We use VGG-13
for SN-GAN experiments and VGG-19 for StyleGAN experiments.

For the rejection sampling (ground truth), the FID score is computed
between $6,000$ images sampled proportional to the probabilities
obtained from classifier and $6,000$ images from the CIFAR-10
dataset. For the outsourced SB the FID score is computed between
$6,000$ samples from the learnt model and $6,000$ real CIFAR-10
samples. All scores are computed only on the images of a specific
class. All other metrics (path KL, mean log-reward, $L^2_2(x_0,
x_1)$, ELBO) are computed using a batch size of 512.

\section{Additional results}\label{appendix:results}
\begin{table}[t] \vspace*{0em} \caption{SB metrics for varying the
    number of time discretisation steps in data-to-data and
    data-to-energy setting with both learnt and fixed variance (Gauss
  $\leftrightarrow$ GMM).} \vspace*{-1em} \resizebox{\linewidth}{!}{
    \begin{tabular}{@{}lllllllll} \toprule Number of steps
      $\rightarrow$ &  \multicolumn{2}{c}{$K=5$} &
      \multicolumn{2}{c}{$K=10$} & \multicolumn{2}{c}{$K=20$} &
      \multicolumn{2}{c}{$K=40$}
      \\ \cmidrule(lr){2-3}\cmidrule(lr){4-5}\cmidrule(lr){6-7}
      \cmidrule(lr){8-9} Algorithm\ $\downarrow$ Metric $\rightarrow$
      & \multicolumn1c{$\mathcal{W}_2^2$} & \multicolumn1c{Path KL} &
      \multicolumn1c{$\mathcal{W}_2^2$} & \multicolumn1c{Path KL} &
      \multicolumn1c{$\mathcal{W}_2^2$} & \multicolumn1c{Path KL} &
      \multicolumn1c{$\mathcal{W}_2^2$} & \multicolumn1c{Path KL}
      \\ \midrule Data-to-data learnt var.     & 0.026\std{0.006}  &
      4.158\std{0.986}  & 0.033\std{0.012}  & 3.127\std{0.981}  &
      0.042\std{0.018}  & 2.840\std{0.668}  & 0.028\std{0.009}  &
      3.070\std{0.666}   \\ Data-to-data fixed var.      &
      0.066\std{0.005}  & 0.988\std{0.031}  & 0.038\std{0.005}  &
      1.825\std{0.073}  & 0.037\std{0.014}  & 2.507\std{0.366}  &
      0.023\std{0.010}  & 2.739\std{0.233}   \\ \midrule
      Data-to-energy learnt var    & 0.023\std{0.010}  &
      3.745\std{0.415}  & 0.022\std{0.008}  & 3.162\std{0.308}  &
      0.028\std{0.015}  & 3.337\std{0.775}  & 0.020\std{0.003}  &
      2.838\std{0.609}   \\ Data-to-energy fixed var     &
      0.063\std{0.006}  & 1.152\std{0.172}  & 0.050\std{0.024}  &
      2.047\std{0.123}  & 0.035\std{0.014}  & 3.069\std{0.226}  &
      0.021\std{0.010}  & 3.360\std{0.150}   \\ \bottomrule
  \end{tabular} } \label{tab:num_steps_ablation}
\end{table}

In addition to \Cref{fig:num_steps_ablation} we also provide the
metrics for the ablation in \Cref{tab:num_steps_ablation}. We use the
same architecture and hyperparameters as in data-to-data experiments
and vary only the number of sampling steps. For the data-to-energy
runs we use a replay buffer with Langevin updates, the off-policy
ratio is set to 0.8 and the backward trajectories are not reused for
the loss computation. $\mathcal{W}^2_2$ is computed using $10,000$
ground truth samples and samples from $\overrightarrow{p}(\tau \mid
x_0)p_0(x_0)$. Path KL is also computed on $10,000$ samples.

\section{Visual examples for outsourced SB}\label{appendix:visual}

\subsection{Uncurated examples}
\begin{figure}[h!]
  \centering
  \begin{tabular}{@{}cc@{}}
    Prior & Posterior \\
    \includegraphics[width=0.4\linewidth]{
      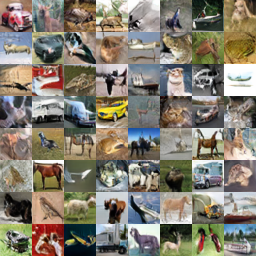
    } &
    \includegraphics[width=0.4\linewidth]{
      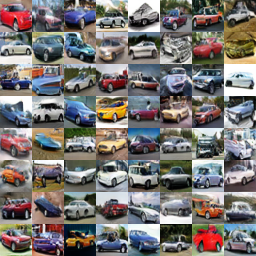
    }
  \end{tabular}
  \caption{Uncurated examples of outsourced SB with SN-GAN for the
  class \textit{cars}.}
\end{figure} 
\begin{figure}[h!]
  \centering

  \begin{tabular}{@{}cc@{}}
    Prior & Posterior \\
    \includegraphics[width=0.4\linewidth]{
      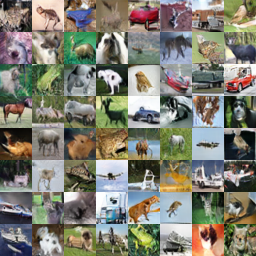
    } &
    \includegraphics[width=0.4\linewidth]{
      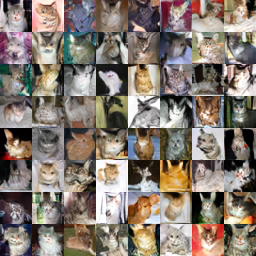
    }
  \end{tabular}
  \caption{Uncurated examples of outsourced SB with SN-GAN for the
  class \textit{cats}.}
\end{figure} 
\begin{figure}[h!]
  \centering
  \begin{tabular}{@{}cc@{}}
    Prior & Posterior \\
    \includegraphics[width=0.4\linewidth]{
      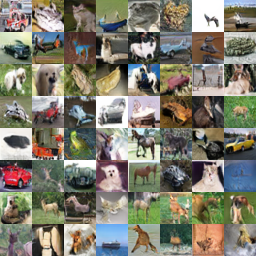
    } &
    \includegraphics[width=0.4\linewidth]{
      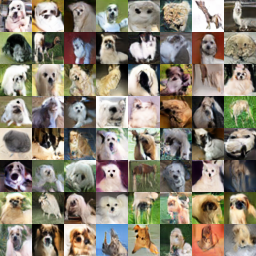
    }
  \end{tabular}
  \caption{Uncurated examples of outsourced SB with SN-GAN for the
  class \textit{dogs}.}
\end{figure} 
\begin{figure}[h!]
  \centering

  \begin{tabular}{@{}cc@{}}
    Prior & Posterior \\
    \includegraphics[width=0.4\linewidth]{
      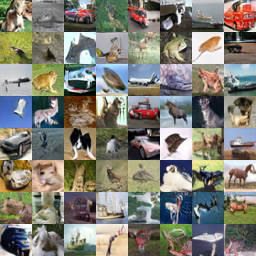
    } &
    \includegraphics[width=0.4\linewidth]{
      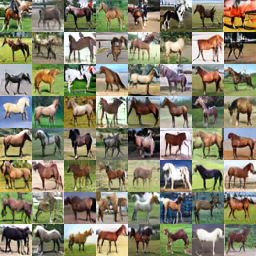
    }
  \end{tabular}
  \caption{Uncurated examples of outsourced SB with SN-GAN for the
  class \textit{horses}.}
\end{figure} 
\begin{figure}[h!]
  \centering

  \begin{tabular}{@{}cc@{}}
    Prior & Posterior \\
    \includegraphics[width=0.4\linewidth]{
      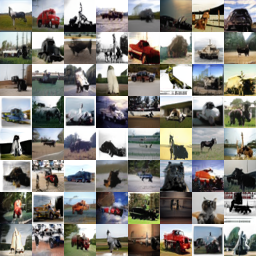
    } &
    \includegraphics[width=0.4\linewidth]{
      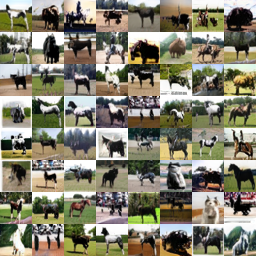
    }
  \end{tabular}
  \caption{Uncurated examples of outsourced SB with StyleGAN for the
  class \textit{horses}.}
\end{figure} 
\begin{figure}[h!]
  \centering

  \begin{tabular}{@{}cc@{}}
    Prior & Posterior \\
    \includegraphics[width=0.4\linewidth]{
      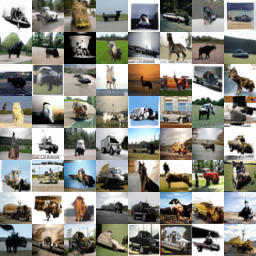
    } &
    \includegraphics[width=0.4\linewidth]{
      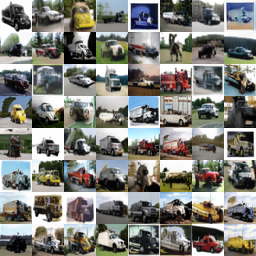
    }
  \end{tabular}
  \caption{Uncurated examples of outsourced SB with StyleGAN for the
  class \textit{trucks}.}
\end{figure} 

\begin{figure}[ht] \centering
  \begin{tabular}{@{}cc@{}} Posterior (Cats) & Posterior (Dogs)
    \\ \includegraphics[width=0.4\linewidth]{
    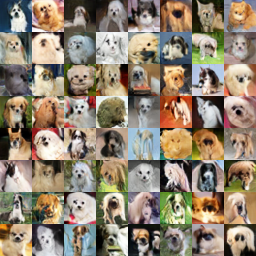 } &
    \includegraphics[width=0.4\linewidth]{
    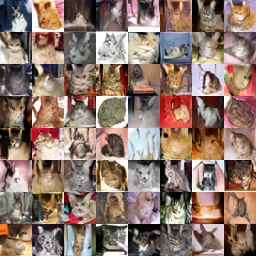 }
  \end{tabular} \caption{ Uncurated examples of outsourced SB with
  SN-GAN between classes \textit{cats} and \textit{dogs}. }
\end{figure} 
\begin{figure}[ht] \centering
  \begin{tabular}{@{}cc@{}} Posterior (Cars) & Posterior (Trucks)
    \\ \includegraphics[width=0.4\linewidth]{
    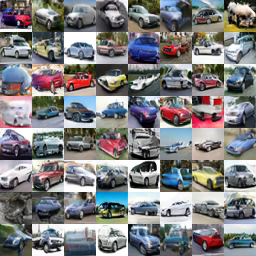 } &
    \includegraphics[width=0.4\linewidth]{
    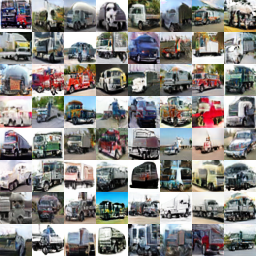 }
  \end{tabular} \caption{ Uncurated examples of outsourced SB  with
  SN-GAN between classes \textit{cars} and \textit{trucks}. }
\end{figure} 
\end{document}